\documentclass{article}

\PassOptionsToPackage{numbers, compress}{natbib}

\usepackage[most]{tcolorbox}

\tcbset{
  csouter/.style={
    enhanced,
    colback=black!2,
    colframe=black!6,
    boxrule=0.3pt,
    arc=4pt,
    left=5pt,right=5pt,top=5pt,bottom=5pt
  },
  csquestion/.style={
    enhanced,
    colback=white,
    colframe=black!18,
    boxrule=0.4pt,
    arc=3pt,
    left=4pt,right=4pt,top=3pt,bottom=3pt,
    fontupper=\small\raggedright
  },
  cscard/.style={
    enhanced,
    colback=white,
    colframe=black!18,
    boxrule=0.4pt,
    arc=3pt,
    left=5pt,right=5pt,top=8pt,bottom=4pt,
    fontupper=\small\raggedright
  }
}

\newtcolorbox{problemBox}{
  colback=gray!10,
  colframe=gray!60,
  boxrule=0.5pt,
  coltitle=white,
  colbacktitle=black,
  fonttitle=\bfseries,
  title={Question},
  arc=3pt,
  left=4pt, right=4pt, top=3pt, bottom=3pt,
  width=\linewidth,
  before upper={\raggedright\small},
}

\newtcolorbox{methodBox}[1]{
  colback=gray!10,
  colframe=gray!60,
  boxrule=0.5pt,
  coltitle=white,
  colbacktitle=black,
  fonttitle=\bfseries,
  title={#1},
  arc=3pt,
  left=4pt, right=4pt, top=3pt, bottom=3pt,
  width=\linewidth,
  before upper={\raggedright\small},
}

\newtcolorbox{questionBox}{
  colback=gray!10,
  colframe=gray!60,
  boxrule=0.5pt,
  coltitle=white,
  colbacktitle=black,
  fonttitle=\bfseries,
  title={Question},
  arc=3pt,
  left=4pt, right=4pt, top=3pt, bottom=3pt,
  width=\linewidth,
  before upper={\raggedright\small}
}

\newtcolorbox{unexploredBox}{
  colback=red!4,
  colframe=red!45!black,
  boxrule=0.5pt,
  coltitle=white,
  colbacktitle=red!55!black,
  fonttitle=\bfseries,
  title={TimeClaw unexplored},
  arc=3pt,
  left=4pt, right=4pt, top=3pt, bottom=3pt,
  width=\linewidth,
  before upper={\raggedright\small}
}

\newtcolorbox{exploredBox}{
  colback=green!4,
  colframe=green!40!black,
  boxrule=0.5pt,
  coltitle=white,
  colbacktitle=green!35!black,
  fonttitle=\bfseries,
  title={TimeClaw explored},
  arc=3pt,
  left=4pt, right=4pt, top=3pt, bottom=3pt,
  width=\linewidth,
  before upper={\raggedright\small}
}

\usepackage[preprint]{neurips_2026}
\usepackage{tikzducks}
\usepackage{graphicx}
\usepackage{subfigure}
\usepackage{booktabs} % for professional tables
\usepackage{pifont}
\usepackage{bm}
\usepackage{makecell}

\usepackage{algorithm}
\usepackage{algorithmicx}
\usepackage[perpage]{footmisc}
\usepackage{fontawesome}
\usepackage{utfsym}

\usepackage[utf8]{inputenc} % allow utf-8 input
\usepackage[T1]{fontenc}    % use 8-bit T1 fonts
\usepackage{url}            % simple URL typesetting
\usepackage{amsfonts}       % blackboard math symbols
\usepackage{nicefrac}       % compact symbols for 1/2, etc.
\usepackage{microtype}      % microtypography
\usepackage{xcolor}         % colors
\usepackage{amsmath}
\usepackage{amssymb}
\usepackage{mathtools}
\usepackage{amsthm}
\usepackage{wrapfig, lipsum}
\usepackage{color, colortbl}
\usepackage{bbm}
\usepackage{subcaption}
\usepackage{multirow}
\usepackage{tikz}
\usepackage{enumitem}

\DeclareMathOperator*{\argmax}{argmax} 
\definecolor{tableofcontent}{HTML}{E63E15}
\definecolor{urlcol}{HTML}{2470D8}
\usepackage[normalem]{ulem}
\useunder{\uline}{\ul}{}
\definecolor{tabblue}{HTML}{5555CC}
\definecolor{brightmaroon}{rgb}{0.76, 0.13, 0.28}

\usepackage[
colorlinks,            
linkcolor=brightmaroon,   
anchorcolor=blue,           
citecolor=tabblue]
{hyperref}

\newcommand{\pgftextcircled}[1]{
    \setbox0=\hbox{#1}%
    \dimen0\wd0%
    \divide\dimen0 by 2%
    \begin{tikzpicture}[baseline=(a.base)]%
        \useasboundingbox (-\the\dimen0,0pt) rectangle (\the\dimen0,1pt);
        \node[circle,draw,outer sep=0pt,inner sep=0.1ex] (a) {#1};
    \end{tikzpicture}
}
\setlist{leftmargin=10mm}
\definecolor{Gray}{gray}{0.9}

\newcommand{\ie}{\textit{i.e., }}
\newcommand{\eg}{\textit{e.g., }}

\usepackage[capitalize,noabbrev]{cleveref}

\usepackage{graphicx}
\usepackage{tikz}
\usepackage{tikzducks}
\theoremstyle{plain}
\newtheorem{theorem}{Theorem}[section]
\newtheorem{proposition}[theorem]{Proposition}

\theoremstyle{definition}

\theoremstyle{remark}

\usepackage{xspace}

\definecolor{darkgreen}{RGB}{0,100,0}

\usepackage[algo2e]{algorithm2e}
\usepackage{extarrows}

\newcommand{\tabref}[1]{Table \ref{#1}}
\newcommand{\secref}[1]{Section \ref{#1}}

\newcommand{\algoref}[1]{Algorithm \ref{#1}}

\def\eg{\emph{e.g., }} 
\def\ie{\emph{i.e., }}

\definecolor{tabblue}{HTML}{5555CC}

\definecolor{tabblue}{HTML}{5555CC}
\definecolor{brightmaroon}{rgb}{0.76, 0.13, 0.28}

\title{
\begin{minipage}{0.1\textwidth}
    \centering
    \includegraphics[width=\linewidth]{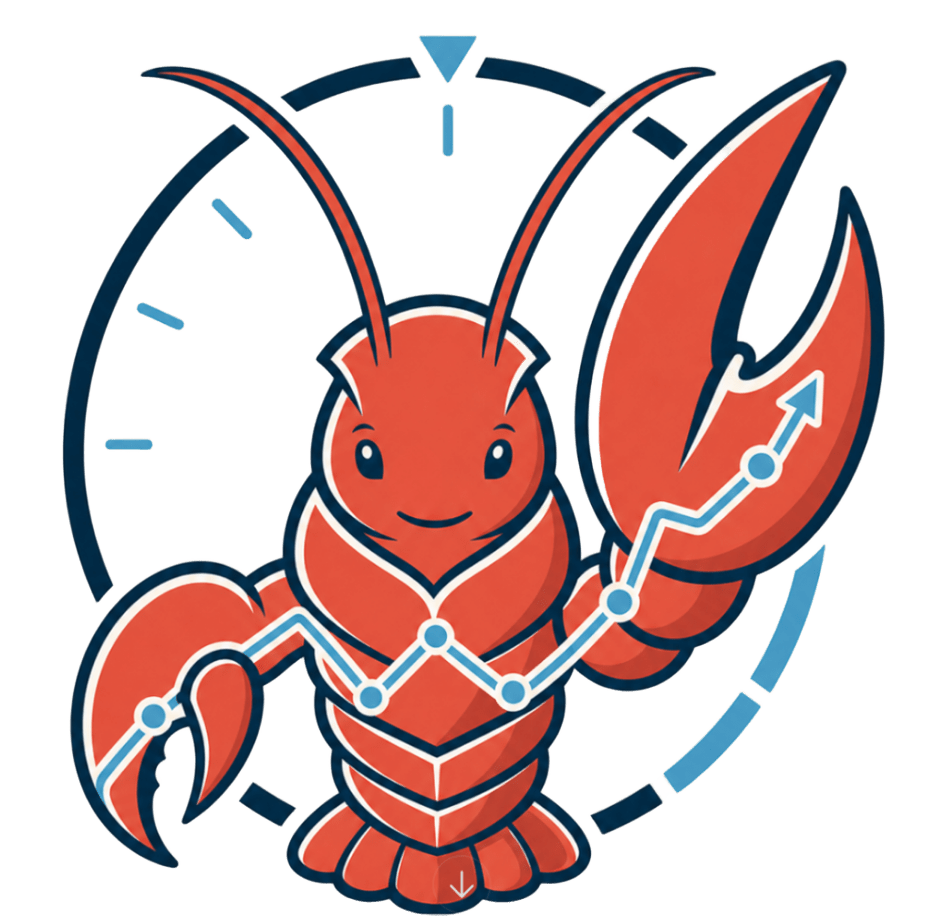}
\end{minipage}
\hspace{0.02\textwidth}
\begin{minipage}{0.86\textwidth}
    \centering
    TimeClaw: A Time-Series AI Agent with Exploratory Execution Learning
\end{minipage}
}
\author{%
  \textbf{Hangchen Liu}$^{1}$,
  \textbf{Dongyuan Li}$^{1}$\thanks{Corresponding author.},
  \textbf{Renhe Jiang}$^{1}$,
  \textbf{Jiewen Deng}$^{2}$,
  \textbf{Weiwei Ye}$^{1}$,
  \textbf{Yoshihide Sekimoto}$^{1}$ \\[0.4em]
  $^{1}$The University of Tokyo,
  $^{2}$Southern University of Science and Technology \\[0.4em]
  \texttt{\{liuhc3,lidy,jiangrh,wwye,sekimoto\}@csis.u-tokyo.ac.jp},
  \texttt{dengjw1@outlook.com} \\[0.4em]
}

\begin{document}

\maketitle

\begin{abstract}
Time series analysis underpins forecasting, monitoring, and decision making in domains such as finance and weather, where solving a task often requires both numerical accuracy and contextual reasoning. Recent progress has moved from specialized neural predictors to approaches built on LLMs and foundation models that can reason over time series inputs and use external tools. However, most such systems remain execution-centric: they focus on solving the current instance but learn little from exploratory execution. This is especially limiting in verifiable numeric settings, where multiple candidate executions and tool-use procedures may all be task-valid yet differ sharply in quantitative quality, and where early success can trigger tool-prior collapse that suppresses further exploration. To address this limitation, we present \textbf{TimeClaw}, an exploratory execution learning framework that turns exploratory execution into reusable hierarchical distilled experience through a four-stage loop: Explore, Compare, Distill, and Reinject. TimeClaw combines metric-supervised exploratory execution learning, task-aware tool dropout, and hierarchical distilled experience for inference-time reinjection, while keeping the base model frozen and avoiding online test-time adaptation. In an MTBench-aligned evaluation with 17 tasks that span finance and weather prediction and reasoning tasks, TimeClaw delivers consistent gains over the baselines. These results suggest that, for scientific systems, the bottleneck is not only execution-time capability, but how exploratory experience is compared, distilled, and reused.

\end{abstract}

% =============================================================================
% TimeClaw — Introduction (NeurIPS 2026 format)
% =============================================================================
% Place this file as \input{introduction} in your main .tex

\section{Introduction}
\label{sec:intro}
\vspace{-5pt}
Large language model (LLM) agents are evolving from one-shot assistants into persistent systems that interact with tools, maintain external state, and support increasingly complex workflows~\cite{he2026trajectbencha,wang2026continuoustime,liu2026spiral,li2026comind,agentbench,webarena,autogen}. This trend is promising for scientific domains such as time-series analysis, where solving a task often requires combining statistical computation, contextual reasoning, and structured decision making~\cite{varambally2025zephyrus,lai2026ustbench,jeon2026stairsformer,zhou2026netarena}. Recent work has made substantial progress on time-series benchmarks and tool-augmented time-series agents~\cite{mtbench,temporalbench,timeart,tsagent,tssci,timecap}. However, most existing time-series agent systems still focus on \emph{execution-time capability}: they improve how an agent solves the current task, but only partially address how experience from exploratory execution should be accumulated and reused across future tasks.

We identify two concrete limitations of the current paradigm: 
\textbf{(i) Completion-only supervision is insufficient for numeric time-series tasks}. In forecasting, indicator prediction, and related numeric tasks, multiple candidate executions can produce task-valid answers, but differ substantially in quantitative quality under metrics such as MAE or RMSE. 
% In such cases, binary notions of success are insufficient for learning: most existing agentic systems evaluate execution outcomes through completion-only supervision, providing only a binary signal indicating whether the task is solved. This collapses quantitatively unequal executions into the same label.
In such cases, completion-only supervision is insufficient because it provides only a binary signal of whether the task is solved, treating executions with different numerical quality as equally successful. 
The key question is not merely whether an agent can complete a task, but whether it can identify, preserve, and reuse \emph{better execution strategies}, including both stronger tool-use procedures and more effective ways of using the selected tools. 
This challenge is especially pronounced in agentic settings: unlike deep learning models that optimize differentiable losses directly, agents cannot backpropagate through discrete tool-use executions. This motivates agent learning in time-series domains to move beyond completion-only supervision toward \textbf{execution-quality-aware learning}, where the learning signal reflects quantitative differences among valid candidate executions~\cite{timeart,tsagent,tssci,castr1,timexl,yeh2025llmagentforecast}. 
\textbf{(ii) Exploratory learning suffers from tool-prior collapse.} 
During exploration, an agent often prematurely develops a strong bias toward a small set of familiar tools that frequently succeed early, then overcommits to them, and suppresses further exploration. We term this failure mode \textbf{tool-prior collapse}: the premature concentration of tool usage on a narrow subset before sufficient comparative evidence has been gathered. Once this happens, execution diversity shrinks, comparative supervision weakens, and distillation into reusable experience becomes less informative. It is important to distinguish this from inference-time tool preference: after learning, an agent may favor certain tools based on evidence, which is a consequence of learning rather than a failure. Tool-prior collapse is an exploration-phase pathology, where entropy in the tool-usage distribution decreases before the agent has explored enough to justify convergence.

To address these limitations, we propose \textbf{TimeClaw}, an exploratory execution learning framework for time-series agents. Rather than treating each task instance as an isolated execution problem, TimeClaw turns exploration-time execution into a reusable source of scientific experience. 
The agent explores multiple candidate executions, compares them under task-valid supervision, distills the resulting experience into reusable hierarchical distilled experience, and reinjects that experience into future problem solving. 
%In this way, execution becomes not only an evaluation target, but also a metric-supervised learning process. 
Crucially, TimeClaw separates learning from inference: learning happens during a dedicated exploration phase, while downstream inference reuses the distilled experience without further online adaptation. 
The framework is built around three key ideas: \textbf{(i) Metric-supervised exploratory execution learning} for numeric and verifiable time-series tasks, enabling the agent to learn not only whether a task is solvable, but which candidate execution strategy achieves better quantitative outcomes; \textbf{(ii) Forced exploration with task-aware tool dropout} to mitigate tool-prior collapse by reducing overreliance on dominant tools and increasing execution diversity; \textbf{(iii) Hierarchical distilled experience} that stores reusable experience outside model weights, including execution-level lessons, tool guidance, and higher-level reusable skills, and reinjects them at inference time via prompt construction. 
In this sense, TimeClaw intersects with tool learning, but goes beyond tool-use correctness by learning from comparative execution quality and reusing distilled experience across tasks~\cite{toolformer,gorilla,toolllm,memskill,asda,skillsd,react,reflexion}.
This positions TimeClaw as a verifiable scientific-domain step from stateless tool use toward persistent, learning-enabled agent systems, where experience accumulation can be measured and interpreted~\cite{openclawrl,cuaskill,clawkeeper}. Overall, our main contributions are summarized as follows:
\begin{enumerate}[leftmargin=1.4em, itemsep=0pt, topsep=0pt, parsep=0pt,partopsep=0pt, label=\textbullet]
% \item We formulate exploratory execution learning for time-series agents and instantiate it in TimeClaw, a learning architecture that separates exploration-time acquisition from inference-time reuse.
% \item We identify execution-quality supervision as a more suitable learning signal than completion-only supervision for verifiable time-series tasks, and operationalize it via metric-supervised learning. 
\item We formulate exploratory execution learning for time-series agents and instantiate it in TimeClaw through {metric-supervised exploratory execution learning}, enabling agents to compare task-valid candidate executions by quantitative quality rather than completion alone (Section~\ref{sec:exploration}).
% \item We identify tool-prior collapse as the premature concentration on familiar tools during exploration before sufficient comparative evidence is gathered, and propose forced exploration with scope-aware tool dropout to improve candidate diversity and supervision quality.
\item We identify {tool-prior collapse} as an exploration-time failure mode in exploratory tool learning and propose {forced exploration with task-aware tool dropout} to preserve candidate diversity and improve supervision quality (Section~\ref{sec:dropout}).
% \item We build an exploratory learning setup for time-series agents and show that hierarchical distilled experience improves downstream reasoning without parameter updates or online adaptation.
\item We design hierarchical distilled experience for inference-time reuse, storing execution lessons, tool guidance, and reusable skills outside model weights to improve downstream reasoning without parameter updates or online adaptation (Section~\ref{sec:distillation}).
\end{enumerate}

% % ── Table 1 ───────────────────────────────────────────────────────────────
% \begin{table}[t]
% \centering
% \caption{Positioning of TimeClaw and related work across four dimensions: tool use, exploratory execution learning, reusable experience distillation, and metric-based execution comparison.}
% \label{tab:positioning}
% \small
% \setlength{\tabcolsep}{5pt}
% \begin{tabular}{@{}lcccc@{}}
% \toprule
% \textbf{Work}
%   & \textbf{Tool Use}
%   & \textbf{Explore Learn.}
%   & \textbf{Distill Reuse}
%   & \textbf{Metric Compare} \\
% \midrule
% TimeART~\cite{timeart}         & \ding{51} & \ding{51} &       &       \\
% TS-Agent~\cite{tsagent}        & \ding{51} &       &       &       \\
% TSci~\cite{tssci}              & \ding{51} &       &       &       \\
% Cast-R1~\cite{castr1}          & \ding{51} & \ding{51} &       &       \\
% ToolLLM~\cite{toolllm}         & \ding{51} &       &       &       \\
% MemSkill~\cite{memskill}       &       &       & \ding{51} &       \\
% ASDA~\cite{asda}               &       &       & \ding{51} &       \\
% Skill-SD~\cite{skillsd}        & \ding{51} & \ding{51} & \ding{51} &       \\
% \midrule
% \textbf{TimeClaw}              & \ding{51} & \ding{51} & \ding{51} & \ding{51} \\
% \bottomrule
% \end{tabular}
% \end{table}

%% ========================================================================
%%  SECTION 2 — RELATED WORK
%% ========================================================================
\vspace{-6pt}
\section{Related Work}
\label{sec:related}
% Recent work related to TimeClaw spans three closely connected directions: time-series agent systems, agent learning through tools and reusable experience artifacts, and persistent open agent runtimes.

% \textbf{Time-series Agent Systems.} Recent work has substantially strengthened time-series agents through tool use, iterative reasoning, forecasting workflow orchestration, and multi-agent collaboration. Among the closest prior systems, TimeART~\cite{timeart} teaches a time-series reasoning model to use tools through expert trajectories. TS-Agent~\cite{tsagent} emphasizes grounded reasoning over raw numeric sequences through iterative evidence gathering, while TimeSeriesScientist~\cite{tssci} and other forecasting-oriented frameworks focus on automating forecasting pipelines and improving end-to-end workflow quality. In parallel, systems such as TS-Debate~\cite{tsdebate} and MAS4TS~\cite{mas4ts} improve performance through debate, modality specialization, or multi-agent decomposition within a single task instance. Cast-R1~\cite{castr1} brings learning-based tool use closer to time-series forecasting by formulating forecasting as sequential decision making with reflective tool use. These approaches primarily optimize execution-time capability. TimeClaw differs in objective: it treats exploratory execution itself as a reusable learning signal, and asks how experience from exploration should be accumulated, distilled, and reused across future tasks.

\vspace{-5pt}
\textbf{Time-series Agent Systems.} Recent work has strengthened time-series agents along three main directions.
First, tool-augmented reasoning systems such as TimeART~\cite{timeart} and TS-Agent~\cite{tsagent,dotiagents} improve how agents reason over raw time-series inputs and use external tools, either through expert tool-use trajectories or iterative evidence gathering.
Second, forecasting-oriented systems such as TimeSeriesScientist~\cite{tssci} automate forecasting workflows and improve end-to-end pipeline quality~\cite{timecap,timexl,flairrts,yeh2025llmagentforecast}.
Third, multi-agent and reflection-based systems such as TS-Debate~\cite{tsdebate}, MAS4TS~\cite{mas4ts}, and Cast-R1~\cite{castr1} improve performance through debate, specialization, decomposition, or reflective tool use within a single task instance~\cite{llmmeetsts,zhang2025competitionnews}.
Despite these advances, existing systems primarily optimize execution-time capability. \textbf{TimeClaw differs in objective}: it treats exploratory execution as a reusable learning signal, and asks how exploration experience should be accumulated, distilled, and reused across future tasks.
By doing so, TimeClaw moves beyond improving one-off execution toward enabling time-series agents to learn reusable strategies from metric-supervised exploration.

\textbf{Agent Learning with Tools and External Experience.}
Recent agent-learning research has increasingly explored how agents can improve across interactions by using tools, retaining external experience, and learning from execution traces~\cite{react,reflexion,voyager,memgpt,autogen}.
Tool-learning methods such as Toolformer~\cite{toolformer}, Gorilla~\cite{gorilla}, and ToolLLM~\cite{toolllm} improve tool invocation, API selection, and action grounding.
Memory- and skill-based methods further store reusable experience outside model weights: MemSkill~\cite{memskill} studies evolving memory skills, ASDA~\cite{asda} distills structured skill files for inference-time use, and Skill-SD~\cite{skillsd} compresses multi-turn trajectories into reusable guidance~\cite{memskill,asda,skillsd,cuaskill}.
Execution-trace and persistent-runtime methods extend this idea by learning from self-generated traces, reward signals, long-lived memory, plugins, and runtime control layers~\cite{openclawrl,cuaskill,clawkeeper,star,rest,clawsafety,safeclawr}.
Together, these works mark a shift from stateless tool use toward agents that retain and reuse experience across interactions. 
In Table~\ref{tab:positioning}, TimeClaw complements these directions but differs in objective and mechanism.
It targets verifiable time-series tasks, where exploratory executions can be compared using task metrics, and turns benchmark execution into reusable learning.
To our knowledge, TimeClaw provides the first time-series agent framework that learns from metric-based comparisons among exploratory executions and distills experience for reuse.

% ── Table 1 ───────────────────────────────────────────────────────────────
\begin{table}[t]
\centering
\caption{Positioning of TimeClaw and related work across four dimensions: tool use, exploratory execution learning, reusable experience distillation, and metric-based execution comparison.}
\label{tab:positioning}
\small
\setlength{\tabcolsep}{3pt}
\resizebox{\linewidth}{!}{
\begin{tabular}{@{}lccccccccc@{}}
\toprule
\textbf{Dimension}
  & \makecell{\textbf{TimeART}\\\cite{timeart}}
  & \makecell{\textbf{TS-Agent}\\\cite{tsagent}}
  & \makecell{\textbf{TSci}\\\cite{tssci}}
  & \makecell{\textbf{Cast-R1}\\\cite{castr1}}
  & \makecell{\textbf{ToolLLM}\\\cite{toolllm}}
  & \makecell{\textbf{MemSkill}\\\cite{memskill}}
  & \makecell{\textbf{ASDA}\\\cite{asda}}
  & \makecell{\textbf{Skill-SD}\\\cite{skillsd}}
& \makecell{\textbf{TimeClaw}\\{Ours}} \\
\midrule
Tool Use        & \textcolor{red}{\ding{51}} & \textcolor{red}{\ding{51}} & \textcolor{red}{\ding{51}} & \textcolor{red}{\ding{51}} & \textcolor{red}{\ding{51}} & \ding{55}      &   \ding{55}    & \textcolor{red}{\ding{51}} & \textcolor{red}{\ding{51}} \\
Explore Learn.  & \textcolor{red}{\ding{51}} &   \ding{55}    &  \ding{55}     & \textcolor{red}{\ding{51}} &   \ding{55}    & \ding{55}      &   \ding{55}    & \textcolor{red}{\ding{51}} & \textcolor{red}{\ding{51}} \\
Distill Reuse   &  \ding{55}     &  \ding{55}     &  \ding{55}     &  \ding{55}     &   \ding{55}    & \textcolor{red}{\ding{51}} & \textcolor{red}{\ding{51}} & \textcolor{red}{\ding{51}} & \textcolor{red}{\ding{51}} \\
Metric Compare  & \ding{55}      &   \ding{55}    &  \ding{55}     &  \ding{55}     &  \ding{55}     &   \ding{55}    &  \ding{55}     &  \ding{55}     & \textcolor{red}{\ding{51}} \\
\bottomrule
\end{tabular}
}
\end{table}

\section{Problem Definition}
\label{sec:problem}

% Each task instance is $x = (z, c, \tau, s)$, where $z$ is the time-series input, $c$ is optional multimodal context, $\tau$ is the task type, and $s$ is the scope for storing reusable experience. The agent explores a set of candidate executions $\Pi(x) = \{\pi_1,\ldots,\pi_K\}$, each a structured trajectory of tool selections, intermediate observations, and a final output $\hat{y}_k = \mathrm{Exec}(x,\pi_k)$. An execution is \textbf{task-valid} if $\hat{y}_k$ satisfies the output contract of $\tau$.

\vspace{-5pt}
Each task instance is denoted by $x=(z,c,\tau,s)$, where $z$ is the time-series input, $c$ denotes optional multimodal context, $\tau$ specifies the task type, and $s$ defines the task under which reusable experience is stored. Given $x$, the agent explores a set of candidate executions $\Pi(x)=\{\pi_1,\ldots,\pi_K\}$, where each $\pi_k$ is a structured trajectory consisting of tool choices, intermediate observations, and a final output $\hat{y}_k=\mathrm{Exec}(x,\pi_k)$. An execution is {task-valid} if its output satisfies the contract specified by $\tau$. For numeric tasks, task validity alone is insufficient because multiple valid executions may differ substantially in quantitative quality. We define execution quality and preferred execution as:
\begin{equation}
q(\pi_k;x)=-\mathcal{L}_{\tau}(\hat{y}_k,y^\star),
\qquad
\pi^\star(x)=\argmax_{\pi_k\in\Pi_{\mathrm{valid}}(x)} q(\pi_k;x),
\end{equation}
where $y^\star$ is the task target, $\mathcal{L}_{\tau}$ is a task-dependent loss such as MAE or RMSE, and $\Pi_{\mathrm{valid}}(x)$ denotes the set of task-valid executions. If no candidate execution is validated, the episode records failure evidence instead of forcing an artificial winner.
This formalization shows why completion-only supervision is insufficient: binary success treats valid executions as equivalent despite different task losses. Since agents cannot backpropagate through tool calls or branching decisions, TimeClaw learns by explicitly comparing candidate executions, which we call {execution-quality supervision}.

% For numeric tasks, multiple executions may be task-valid yet differ in quality. We define execution quality as
% \begin{equation}
% q(\pi_k;\, x) = -\mathcal{L}_{\tau}(\hat{y}_k,\, y^\star),
% \end{equation}
% where $y^\star$ is the task target and $\mathcal{L}_{\tau}$ is task-dependent (\eg MAE, RMSE). The preferred execution strategy is
% \begin{equation}
% \pi^\star(x) = \argmax_{\pi_k \in \Pi_{\mathrm{valid}}(x)} q(\pi_k;\, x).
% \end{equation}
% If no candidate is validated, the episode records failure evidence rather than forcing an artificial winner.

% This formalization captures why completion-only supervision is insufficient: a binary success signal maps all task-valid executions to the same label even when $\mathcal{L}_{\tau}(\hat{y}_a, y^\star) \neq \mathcal{L}_{\tau}(\hat{y}_b, y^\star)$. Since agents cannot backpropagate through tool calls and branching decisions, the only way to learn which execution is better is explicit comparison. TimeClaw therefore learns from \textbf{execution-quality supervision}.

% TimeClaw separates the agent lifecycle into two phases. In the \textbf{exploration phase}, the agent generates and compares candidate executions under task-valid supervision. In the \textbf{inference phase}, the agent reuses distilled experience without online learning. This separation keeps the learning process auditable, reusable, and grounded in verifiable task metrics.

%% ========================================================================
%%  SECTION 4 — METHODOLOGY
%% ========================================================================
\section{Methodology}
\label{sec:method}

\begin{figure}[t]
    \centering
    \includegraphics[width=\linewidth]{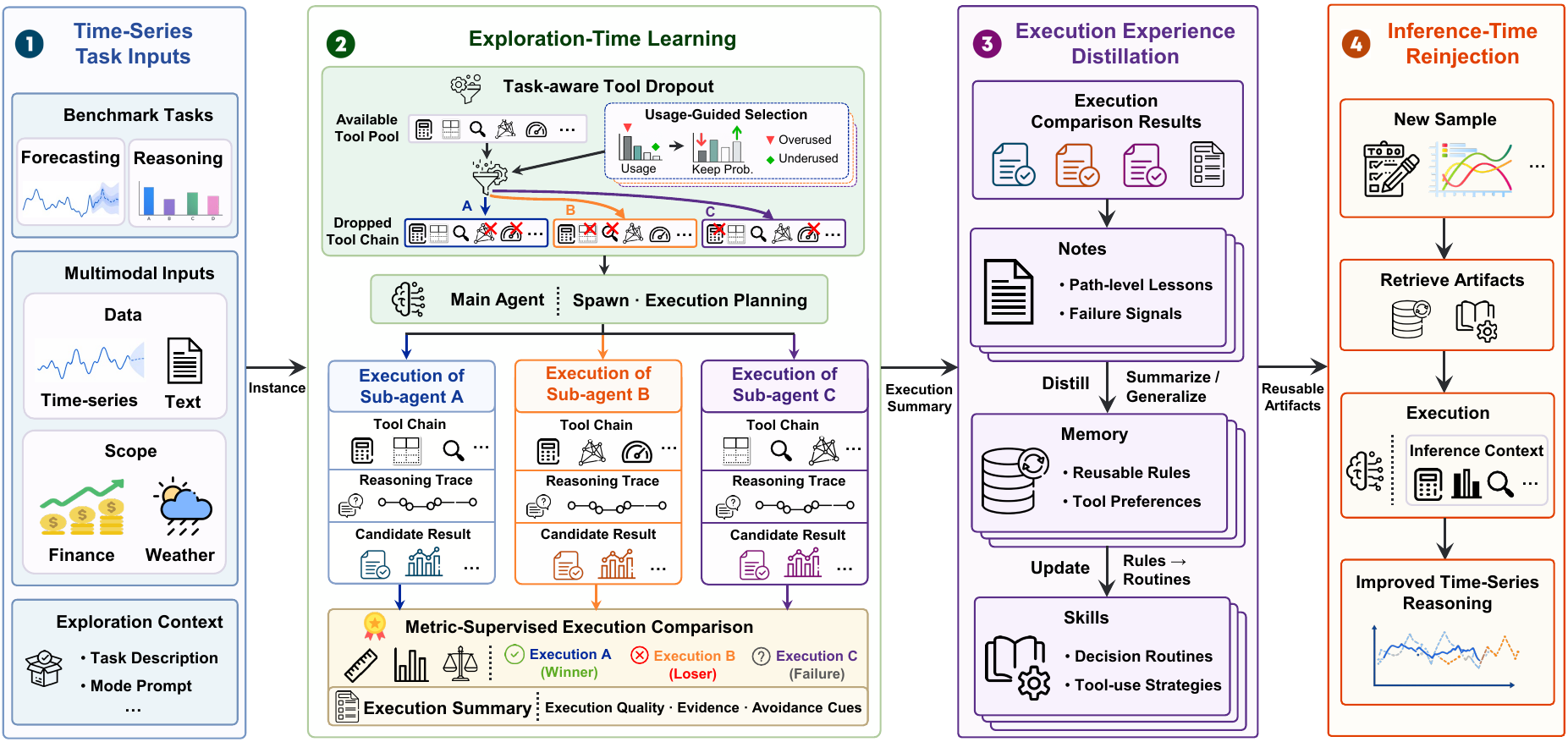}
    \caption{Overview of TimeClaw. The framework follows four stages:  exploration-time learning, metric-supervised execution comparison, execution experience distill, and inference-time reinject.}
    \label{fig:main}
\end{figure}

% \subsection{Framework Overview}
% \label{sec:overview}
\vspace{-5pt}
TimeClaw implements exploratory execution learning as a four-stage loop, \emph{Explore $\to$ Compare $\to$ Distill $\to$ Reinject}, that turns exploration into reusable inference-time experience. 
As shown in Figure~\ref{fig:main}, it first uses a constrained \textbf{exploratory executor} to generate multiple candidate executions and compare task-valid candidates under metric-supervised feedback, so that learning reflects execution quality rather than completion alone (\secref{sec:exploration}). 
It then applies \textbf{task-aware tool dropout} to mitigate tool-prior collapse and preserve diverse candidate executions (\secref{sec:dropout}). 
Finally, it performs \textbf{hierarchical distillation and inference-time reuse}: it distills execution evidence into hierarchical distilled experience and reinjects relevant experience into future prompts without online adaptation or model-parameter updates (\secref{sec:distillation}). 
Thus, exploration accumulates reusable task-local experience, while inference reuses that experience with the base agent kept frozen.

% TimeClaw instantiates exploratory execution learning as a four-stage loop---\emph{Explore\,$\to$\,Compare\,$\to$\,Distill\,$\to$\,Reinject}---whose components are:

% \begin{enumerate}[leftmargin=*,itemsep=2pt]
%     \item \textbf{Explore} (\secref{sec:exploration}): generate multiple candidate executions for the same task via constrained sub-agent branching.
%     \item \textbf{Compare} (\secref{sec:exploration}): evaluate task-valid candidates under metric-supervised comparison to identify the preferred execution strategy.
%     \item \textbf{Distill} (\secref{sec:distillation}): compress execution evidence into reusable hierarchical distilled experience.
%     \item \textbf{Reinject} (\secref{sec:distillation}): inject distilled experience into future inference prompts without online adaptation.
% \end{enumerate}

% Learning deposits all outcomes into reusable distilled experience rather than model parameters. The experience hierarchy (\tabref{tab:artifacts}) contains several layers with distinct semantic roles, from a fixed identity layer (\textsc{Soul}) to scope-local operational procedures (\textsc{Skills}). Scope-aware tool dropout (\secref{sec:dropout}) operates during exploration to mitigate tool-prior collapse.

\subsection{Exploration-Time Agent Learning}
\label{sec:exploration}

The exploration phase is implemented by a constrained executor that turns autonomous execution into explicit exploratory execution learning.  
In exploration mode, the main agent can access exploration-only orchestration tools, including \texttt{spawn\_subagent} for branch generation and \texttt{evaluate\_*} tools for post-hoc comparison against ground truth. 
Unlike ordinary inference, exploration must satisfy an exploration contract: the agent should generate at least two task-valid and distinct candidate executions, compare them before termination, and finish only by producing a \texttt{learning\_summary}. 
When prior experience already exists for the task, the first exploration round must include both a prior-guided candidate and an alternative candidate, to prevent reuse suppress exploration (Appendix~\ref{alg:learn}).

The \texttt{learning\_summary} is not treated as the final task answer. 
Instead, it records the selected execution strategy, the decisive evidence behind the selection, and reusable lessons for future tasks. 
For an exploration instance $x$, TimeClaw first constructs a sample-level learning record:
\begin{equation}
n_x = \mathrm{Summarize}\!\bigl(x,\,\Pi(x),\, \{q(\pi_k;x)\}_{k=1}^{K},\, \pi^\star(x)\bigr),
\end{equation}
where $n_x$ captures the strongest validated execution, decisive evidence, and failure signals from weaker candidates. 
Because raw reflective text may contain task-answer leakage or orchestration-specific wording, TimeClaw rewrites it into reusable execution evidence:
\begin{equation}
e_x = \mathrm{Clean}(n_x),
\end{equation}
where $\mathrm{Clean}(\cdot)$ removes answer leakage, framework-control language, and non-transferable execution details.
The cleaned evidence then updates the reusable experience state for task $s$:
\begin{equation}
\mathcal{A}_{s}^{(m+1)} = \mathrm{Distill}\,\!\bigl(\mathcal{A}_{s}^{(m)},\, e_x\bigr),
\end{equation}
where $\mathcal{A}_{s}$ denotes the current reusable experience state for task $s$. 
The update target is not a model parameter vector, but a structured experience state containing notes, reusable rules, tool guidance, and operational procedures. 
To keep candidate executions semantically comparable, TimeClaw tracks intermediate tool outputs as typed artifacts with coordinate provenance, ensuring that downstream operations are conditioned on the current experience state rather than blindly reusing indices from the original series. 
This is especially important for sliced sequences, derived windows, and forecast artifacts, where index meaning depends on the transformation history.

When the executor cannot produce the minimum required number of task-valid candidates, the system does not discard the episode entirely. If exactly one valid candidate exists, the episode records single-execution evidence without comparative signal. If no valid candidate exists, the episode records failure evidence that can still inform future tool guidance. This fallback ensures that even unsuccessful exploration contributes to the learning process, albeit with weaker signal. The full learning and inference protocols are summarized in \algoref{alg:learn} and \algoref{alg:infer} in Appendix~\ref{Algorithms}.

\subsection{Tool-Prior Collapse and Task-Aware Tool Dropout}
\label{sec:dropout}

A major obstacle to exploratory tool learning is that the agent prematurely develops strong preferences for a small subset of familiar tools. We term this phenomenon \textbf{tool-prior collapse}, where historically frequent or often-successful tools dominate future branch generation before sufficient comparative evidence has been gathered, thereby reducing execution diversity and weakening subsequent comparison and distillation. 
To our knowledge, prior time-series agent work has not explicitly named and operationalized this exploration-time failure mode. 
This phenomenon differs from inference-time tool preference: after learning, favoring certain tools based on accumulated evidence is a consequence of learning rather than a failure. 
By contrast, tool-prior collapse occurs during exploration, where early concentration of tool usage signals premature convergence rather than informed narrowing.

Let $n_i(s)$ denote the historical usage count of tool $i$ under task $s$, and let $p_i(s)$ be the normalized usage distribution over competing tools in that task:
\begin{equation}
p_i(s)=\frac{n_i(s)}{\sum_j n_j(s)}, 
\qquad
\mathcal{H}(s) = - \sum_i p_i(s)\log p_i(s).
\end{equation}
A decreasing $\mathcal{H}(s)$ indicates that tool usage is becoming concentrated on a smaller support. 
In this regime, candidate executions become structurally narrower, and execution comparison becomes less informative because different branches may rely on the same dominant tool family.

TimeClaw mitigates this problem with \textbf{task-aware tool dropout}, which operates during \emph{exploration-time sub-agent tool exposure}. 
For each task, TimeClaw tracks historical tool usage and assigns each candidate tool a keep probability relative to other competing tools in the same task. 
Frequently used tools receive lower keep probabilities, while underused tools remain more visible. 
Task-critical tools and explicitly hinted tools are protected through a \texttt{never\_drop} rule. 
The goal is not generic regularization, but preserving branch diversity so that the agent can generate meaningfully different candidate executions instead of repeatedly collapsing onto the same familiar execution pattern.

The keep probability is a frequency-aware rule anchored at the least-explored tool. 
For each non-protected tool $i$ in task $s$, TimeClaw compares its usage count $n_i(s)$ with the minimum usage count $n_{\min}(s)$ among competing non-protected tools, and defines
\begin{equation}
\label{eq:keep}
p_{\mathrm{keep}}(i \mid s)=
\left(\frac{1+n_{\min}(s)}{1+n_i(s)}\right)^{\!\alpha},
\end{equation}
where $\alpha>0$ controls the strength of reweighting. 
This power-law form gives $p_{\mathrm{keep}}(i\mid s)=1$ to tools at the cold edge of the task, \ie $n_i(s)=n_{\min}(s)$, and increasingly suppresses tools that have been used disproportionately more often. 
Protected tools bypass dropout altogether:
\begin{equation}
p_{\mathrm{keep}}(i \mid s)=1 \quad \text{for } i \in \mathcal{T}_{\mathrm{protected}}(s).
\end{equation}

\begin{proposition}[Monotonicity]
\label{prop:mono}
For non-protected tools $i$ and $j$ in the same task $s$, if $n_i(s) > n_j(s)$, then $
p_{\mathrm{keep}}(i \mid s) < p_{\mathrm{keep}}(j \mid s)
\quad \text{for } \alpha > 0. $
\end{proposition}

By Proposition~\ref{prop:mono}, \textbf{task-aware tool dropout} suppresses historically dominant tools more strongly than less-used ones, while tools at the cold edge of the task remain visible by construction.
This makes the mechanism a targeted anti-collapse intervention rather than a generic regularizer, preserving candidate diversity and yielding more informative execution-quality supervision.

\begin{table}[h!]
\centering
\caption{Hierarchical distilled experience in TimeClaw.}
\label{tab:artifacts}
\small
\setlength{\tabcolsep}{15pt}
\begin{tabular}{@{}llll@{}}
\toprule
\textbf{Layer} & \textbf{Injection} & \textbf{Content} & \textbf{Function} \\
\midrule
Soul    & System prompt  & Fixed identity \& behavior rules       & Stable behavioral anchor \\
Notes   & Not injected   & Sample-level experience records         & Distillation source \\
Memory  & System prompt  & Structured reusable rules (9-tuple)     & Primary runtime guidance \\
Tools   & User prompt    & Tool-specific usage advice              & Per-tool best practices \\
Skills  & User prompt    & task-local SOPs                        & Reusable task procedures \\
\bottomrule
\end{tabular}
\end{table}

\subsection{Hierarchical Distillation, Conflict-Aware Memory, and Inference-Time Reuse}
\label{sec:distillation}

TimeClaw externalizes learning into the hierarchical distilled experience described in \tabref{tab:artifacts}, rather than updating model parameters. 
At the sample level, \textsc{Notes} preserve raw exploration evidence, but they are not the main runtime memory surface. 
Instead, they serve as distillation sources for more stable task-level experience layers. 
The key runtime layer is \textsc{Memory}, whose entries are structured reusable rules rather than free-form reflections. 
A memory rule is represented as a 9-tuple:
\begin{equation}
\label{eq:rule}
r = (\kappa,\,\sigma,\,\chi,\,\mathcal{T}^{+},\,\mathcal{T}^{-},\,\rho,\,\mathcal{E},\,c,\,\iota),
\end{equation}
where $\kappa$ is the rule kind (\eg tool preference, condition-action, avoidance), $\sigma$ is a concise summary, $\chi$ specifies applicability conditions (\eg  task subtypes), $\mathcal{T}^{+}$ and $\mathcal{T}^{-}$ are preferred and avoided tool sets, $\rho$ is the rationale, $\mathcal{E}$ is supporting evidence from exploration episodes, $c\in[0,1]$ is confidence, and $\iota\in\{0,1\}$ indicates whether the rule is currently safe to inject into the prompt.

This structured representation allows TimeClaw to update memory by conflict-aware distillation rather than naive appending. 
Given cleaned evidence $e_x$, the task-local memory state is updated as:
\begin{equation}
\mathcal{M}_{s}^{(m+1)} = \mathrm{Update}\,\!\bigl(\mathcal{M}_{s}^{(m)},\, e_x\bigr),
\end{equation}
where $\mathrm{Update}$ may append a new rule, merge near-duplicate evidence, strengthen a supported rule, or register a conflict relation between competing rules. 
If a rule is not yet stable enough for prompt-time reuse, it can remain stored while being marked non-injectable through $\iota=0$. 
This separation between storage and injection lets TimeClaw preserve long-horizon experience without forcing every intermediate belief into the runtime prompt. The same distillation principle extends beyond \textsc{Memory}. 
\textsc{Tools} distills tool-specific usage advice from repeated execution evidence, while \textsc{Skills} compresses stronger task-local procedures into reusable SOP-like guidance. 
To make learning auditable over time, TimeClaw also records historical snapshots of \textsc{Notes}, \textsc{Memory}, tool notes, and skill text. 
These snapshots provide a long-term trace of how reusable experience evolves, rather than reducing learning to the latest prompt state.

During inference, TimeClaw reuses these artifacts without online learning. 
For a new instance $x'$ under task $s$, inference is written as:
\begin{equation}
\hat{y} = f_{\text{agent}}\,\!\bigl(x';\, \mathrm{Retrieve}(\mathcal{A}_{s}),\, \mathcal{T}(x')\bigr),
\end{equation}
where $\mathrm{Retrieve}(\mathcal{A}_{s})$ returns relevant reusable experience for the task, and $\mathcal{T}(x')$ is the runtime-compatible visible tool set. 
In the present implementation, retrieval is filtered by task, applicability, and injection status. The prompt-eligible memory subset is:
\begin{equation}
\mathcal{R}_{s}(x')=
\bigl\{r \in \mathcal{M}_{s} : \iota(r)=1,\; \mathrm{Match}(\chi(r),\, \phi(x'))=1\bigr\},
\end{equation}
where $\phi(x')$ denotes the sample fingerprint and profiling view of the current instance. Prompt construction follows a fixed order: it first provides the official task description, task boundary, and output contract; then adds the sample fingerprint and profiling signals; and finally reinjects the learned artifact layers. 
In the present implementation, \textsc{Soul} and task-local \textsc{Memory} are injected at the system level, while tool notes and \textsc{Skills} are injected as reusable support guidance. 
The visible tool catalog is further filtered by modality and runtime compatibility. The exploration-test difference is crucial. Exploration exposes exploration-only tools such as \texttt{spawn\_subagent} and \texttt{evaluate\_*} because the goal is to compare candidate executions. Inference removes these exploration-only tools and retains only the tools needed to solve the task itself. Thus, the runtime agent is not continuing to learn; it is solving the task under a prompt enriched by previously distilled experience. The primary reuse mechanism is best understood as \emph{task-local experience reinjection}.

% This train-test separation is crucial. 
% Training exposes exploration-only tools such as \texttt{spawn\_subagent} and \texttt{evaluate\_*} because the goal is to generate and compare candidate executions. 
% Inference removes these training-only tools and retains only the tools needed to solve the task. 
% Thus, the runtime agent does not continue learning online; it solves the task with a prompt enriched by previously distilled experience. 
% The primary reuse mechanism is therefore \emph{scope-local experience reinjection}.

% Prompt construction first assembles an official task description, then provides the task boundary and output contract, then injects the sample fingerprint and profiling signals, and finally reinjects the learned artifact layers. In the present implementation, \textsc{Soul} and scope-local \textsc{Memory} are injected at the system level, while tool notes and skills are injected as reusable support guidance; the visible tool catalog is further filtered by modality and runtime compatibility.

\section{Experiments}

% Auto-generated from optimizer/overleaf/generate_paper30_overleaf_tables.py
% New definition: this file uses the 30-metric paper-table setting only.
% New definition: TimeClaw refers to the merged whole-row selection result.
% New definition: GPT-5.4 refers to the softened two-run whole-row selection result.
% New definition: added MTBench paper baselines here are the fully covered cross-table baselines GPT-4o, Gemini, Claude, and DeepSeek.

\subsection{Experimental Setup}
\vspace{-5pt}
\textbf{Datasets.} We evaluate on a fixed 17-task MTBench evaluation suite~\cite{mtbench,temporalbench}, using official prompts, task contracts, and task-specific label spaces. The suite covers finance and weather forecasting, indicator prediction, trend classification, correlation reasoning, and MCQA. We report 30 paper metrics aggregated over these 17 tasks, following the evaluation task defined in this paper. In addition to the official evaluation suite, TimeClaw performs exploration and experience distillation on a separately generated MTBench-style corpus whose task schemas and contracts are aligned with MTBench while keeping the samples disjoint from the evaluation set. Further details regarding both the evaluation suite and the exploration corpus are provided in appendix~\ref{app:data:learning}.

\textbf{Baselines.} We evaluate against two groups of baselines: closed or general-purpose LLMs, including GPT-4o, Gemini, Claude, DeepSeek, and GPT-5.4, and open models, including Qwen-3, Qwen-3M, Llama-3, and Llama-3M where available. We emphasize GPT-5.4 as the strongest direct baseline while reporting all baselines in our experiments.

% \textbf{Metrics.}
% We organize evaluation metrics in the same way as the main results tables:
% \begin{itemize}[leftmargin=1.5em]
%     \item \textbf{Core numeric prediction and forecasting}: finance prediction is reported with MAE and MAPE; MACD prediction is reported with MSE; weather forecasting is reported with MSE and MAE.
%     \item \textbf{Indicator prediction}: weather-indicator prediction is reported separately with max/min/diff under both MSE and MAE.
%     \item \textbf{Reasoning tasks}: trend classification, correlation reasoning, and MCQA are reported with accuracy; for finance trend and correlation, we report both the finer-grained 5-way metric and the coarser 3-way metric.
% \end{itemize}
% The 30 reported metrics are the task-level paper metrics induced by this 17-task suite. We use them for all main tables and comparisons, together with per-task breakdowns where needed.

\textbf{Metrics.} We report task-specific metrics as follows: (i) core numeric prediction use MAE/MAPE for finance prediction, MSE for MACD prediction, and MSE/MAE for weather forecasting; (ii) weather-indicator prediction is evaluated separately on max, min, and diff values under both MSE and MAE; and (iii) reasoning tasks, including trend classification, correlation reasoning, and MCQA, use accuracy, with finance trend and correlation further reported under both 5-way and 3-way settings. 
% Together, these define 30 task-level metrics over the 17-task suite, which are used for all main comparisons and per-task breakdowns.

\textbf{Implementation.} All methods use the same LLM backbones whenever possible. We keep MTBench prompts and parsing contracts fixed, so the comparison focuses on learning protocol and agent design rather than prompt engineering. For TimeClaw, exploration uses a disjoint MTBench corpus to build hierarchical distilled experience with exploration-only branch generation and ground-truth comparison, while test-time evaluation allows experience reuse without online adaptation~\cite{chronos,moirai,timesfm,timellm,moment,lagllama,liu2025sundial,liu2024timer,wu2025aurora}.
% All methods use the same base LLM backbone whenever possible. We keep the official MTBench prompts and parsing contracts fixed across methods, so the comparison focuses on learning protocol and agent design rather than prompt engineering. For TimeClaw, the only exploration-time capabilities unavailable at test time are the exploration-only branch-generation and ground-truth comparison operations used to build external execution experience. TimeClaw learns on an independently constructed MTBench-compatible exploration corpus whose schemas and task contracts are aligned with MTBench but whose samples are disjoint from the official benchmark. No online adaptation is allowed at test time. During evaluation, the agent only reuses the hierarchical distilled experience produced during the exploration phase.

% Hereafter, we design the experiments to answer four questions: \textbf{RQ1.} Does TimeClaw improve benchmark-grounded time-series reasoning on MTBench? 
% \textbf{RQ2.}  Does metric-supervised exploratory execution learning matter beyond completion-only or single-path execution? 
% \textbf{RQ3.}  Does task-aware tool dropout mitigate tool-prior collapse and improve transfer? 
% \textbf{RQ4.}  Which layers of hierarchical distilled experience are responsible for downstream gains?

\begin{table*}[h]
\centering
\small
\setlength{\tabcolsep}{11pt}
\renewcommand{\arraystretch}{1.08}
\caption{Finance prediction results under the 30-metric paper-table setting.}
\vspace{-6pt}
\label{tab:finance_prediction}
\begin{tabular}{lcccccc}
\toprule[1.1pt]
\multicolumn{1}{l}{Model} & \multicolumn{2}{c}{Stock Forecast (7D)} & \multicolumn{2}{c}{Stock Forecast (30D)} & \multicolumn{1}{c}{MACD (7D)} & \multicolumn{1}{c}{MACD (30D)} \\
\cmidrule(lr){2-3}\cmidrule(lr){4-5}\cmidrule(lr){6-6}\cmidrule(lr){7-7}
 & MAE & MAPE & MAE & MAPE & MSE & MSE \\
\midrule[0.8pt]
GPT-4o & \underline{1.596} & 2.544 & 2.338 & 3.520 & 0.365 & 0.897 \\
Gemini & 1.628 & 3.513 & 2.432 & 3.268 & 0.384 & 0.975 \\
% Claude & 1.422 & 2.098 & 2.065 & 2.847 & 0.373 & 1.171 \\
DeepSeek & 1.720 & 2.135 & \underline{2.134} & 3.305 & \underline{0.352} & 1.072 \\
% OpenAI-o3 & 0.929 & 1.324 & 1.704 & 2.231 & 0.246 & 0.586 \\
Qwen-3 & 2.528 & 4.607 & 3.943 & 14.101 & 1.088 & 1.621 \\
Qwen-3M & 2.607 & 4.610 & 4.125 & 16.364 & 1.072 & 1.527 \\
Llama-3 & 2.653 & 4.618 & 3.780 & 11.209 & 0.448 & \textbf{0.812} \\
Llama-3M & 2.683 & 4.666 & 3.777 & 11.010 & 0.586 & \underline{0.835} \\
% GPT-5.1 & 1.199 & 1.546 & 2.052 & 2.606 & 0.313 & 1.184 \\
% GPT-5.1M & 1.453 & 1.762 & 2.199 & 2.721 & 0.310 & 1.144 \\
GPT-5.4 & 1.660 & \underline{1.843} & 3.570 & \underline{3.021} & 0.411 & 5.336 \\
TimeClaw & \textbf{1.341} & \textbf{1.835} & \textbf{2.032} & \textbf{2.977} & \textbf{0.201} & 1.111 \\
\bottomrule[1.1pt]
\end{tabular}
\vspace{-10pt}
\end{table*}

\begin{table}[h]
\centering
\small
\setlength{\tabcolsep}{3.2pt}
\renewcommand{\arraystretch}{1.08}
\caption{Weather prediction results under the 30-metric paper-table setting.}
\label{tab:weather_prediction}
\resizebox{\columnwidth}{!}{%
\begin{tabular}{lcccccccccc}
\toprule[1.1pt]
\multicolumn{1}{l}{Model} & \multicolumn{2}{c}{Forecast (7D)} & \multicolumn{2}{c}{Forecast (14D)} & \multicolumn{6}{c}{Weather Indicator} \\
\cmidrule(lr){2-3}\cmidrule(lr){4-5}\cmidrule(lr){6-11}
 & MSE & MAE & MSE & MAE & Max MSE & Max MAE & Min MSE & Min MAE & Diff MSE & Diff MAE \\
\midrule[0.8pt]
GPT-4o   & 17.550 & 3.110 & 40.430 & 4.490 & 26.030 & 3.020 & 15.390 & 2.760 & 18.840 & 3.200 \\
Gemini   & 24.310 & 3.670 & 29.470 & 4.030 & 15.790 & 2.940 & 16.270 & 2.930 & 23.210 & 3.630 \\
DeepSeek & 29.380 & 4.040 & 101.28 & 6.610 & 32.820 & 4.380 & 17.250 & 3.050 & 44.990 & 5.240 \\
Qwen-3   & 22.640 & 3.630 & 24.810 & 3.850 & 15.790 & 2.940 & 13.910 & 2.750 & 19.430 & 3.340 \\
Qwen-3M  & 22.210 & 3.610 & 24.810 & 3.840 & 13.530 & 2.710 & 15.710 & 2.960 & 18.340 & 3.270 \\
Llama-3  & 20.470 & 3.540 & 35.430 & 4.670 & 12.540 & \underline{2.570} & 12.680 & 2.630 & \underline{15.100} & \underline{2.900} \\
Llama-3M & 21.610 & 3.560 & 33.910 & 4.530 & \underline{12.100} & 2.590 & 13.000 & 2.710 & 15.700 & 2.930 \\
GPT-5.4  & \underline{14.197} & \underline{2.774} & \underline{20.984} & \underline{3.470} & 13.669 & 2.858 & \underline{10.530} & \underline{2.386} & 15.415 & 2.957 \\
TimeClaw & \textbf{10.719} & \textbf{2.442} & \textbf{13.525} & \textbf{2.744} & \textbf{10.898} & \textbf{2.371} & \textbf{10.507} & \textbf{2.292} & \textbf{13.131} & \textbf{2.623} \\
\bottomrule[1.1pt]
\end{tabular}%
}
\end{table}

\begin{table*}[t]
\centering
\footnotesize
\setlength{\tabcolsep}{4.5pt}
\renewcommand{\arraystretch}{1.06}
\caption{Finance results on trend classification, MCQA, and news-stock correlation.} %under the 30-metric setting.}
\label{tab:finance_all}
\begin{tabular}{lcccccccccc}
\toprule[1.1pt]
\multicolumn{1}{l}{Model}
& \multicolumn{2}{c}{Trend (7D)}
& \multicolumn{2}{c}{Trend (30D)}
& \multicolumn{1}{c}{MCQA (7D)}
& \multicolumn{1}{c}{MCQA (30D)}
& \multicolumn{2}{c}{Corr. (7D)}
& \multicolumn{2}{c}{Corr. (30D)} \\
\cmidrule(lr){2-3}
\cmidrule(lr){4-5}
\cmidrule(lr){6-6}
\cmidrule(lr){7-7}
\cmidrule(lr){8-9}
\cmidrule(lr){10-11}
& Acc.-5 & Acc.-3
& Acc.-5 & Acc.-3
& Acc.
& Acc.
& Acc.-5 & Acc.-3
& Acc.-5 & Acc.-3 \\
\midrule[0.8pt]
GPT-4o
& 0.365 & 0.428
& 0.306 & 0.474
& 0.651
& 0.528
& 0.310 & 0.536
& 0.346 & 0.576 \\
Gemini
& 0.415 & 0.473
& 0.297 & 0.449
& 0.636
& 0.503
& 0.264 & 0.518
& \underline{0.348} & \textbf{0.596} \\
Claude
& 0.334 & 0.449
& 0.317 & 0.521
& 0.756
& 0.611
& 0.290 & 0.504
& 0.343 & \underline{0.579} \\
DeepSeek
& 0.356 & 0.451
& 0.296 & 0.483
& 0.776
& 0.693
& 0.271 & 0.500
& \textbf{0.350} & 0.575 \\
GPT-5.4
& \underline{0.569} & \underline{0.601}
& \underline{0.410} & \underline{0.526}
& \underline{0.821}
& \underline{0.791}
& 0.259 & 0.448
& 0.250 & 0.448 \\
TimeClaw
& \textbf{0.696} & \textbf{0.704}
& \textbf{0.514} & \textbf{0.602}
& \textbf{0.868}
& \textbf{0.795}
& \textbf{0.348} & \textbf{0.580}
& 0.331 & 0.554 \\
\bottomrule[1.1pt]
\end{tabular}
\end{table*}

% \begin{table*}[t]
% \centering
% \footnotesize
% \setlength{\tabcolsep}{4.8pt}
% \renewcommand{\arraystretch}{1.08}
% \caption{Weather reasoning results under the 30-metric paper-table setting.}
% \label{tab:weather_reasoning}
% \begin{tabular}{lcccc}
% \toprule[1.1pt]
% \multicolumn{1}{l}{Model} & \multicolumn{1}{c}{Past Trend} & \multicolumn{1}{c}{Future Trend} & \multicolumn{1}{c}{MCQA (Short)} & \multicolumn{1}{c}{MCQA (Long)} \\
% \cmidrule(lr){2-2}\cmidrule(lr){3-3}\cmidrule(lr){4-4}\cmidrule(lr){5-5}
%  & Acc. & Acc. & Acc. & Acc. \\
% \midrule[0.8pt]
% GPT-4o & 0.664 & 0.435 & 0.417 & 0.448 \\
% Gemini & 0.600 & 0.518 & 0.434 & 0.540 \\
% % Claude & 0.598 & 0.569 & 0.518 & 0.512 \\
% DeepSeek & 0.568 & 0.252 & 0.467 & 0.573 \\
% Qwen-3 & 0.375 & 0.447 & 0.578 & 0.562 \\
% Llama-3 & 0.473 & 0.197 & \underline{0.605} & \underline{0.626} \\
% GPT-5.4 & \underline{0.980} & \underline{0.521} & 0.567 & 0.580 \\
% TimeClaw & \textbf{0.994} & \textbf{0.525} & \textbf{0.616} & \textbf{0.664} \\
% \bottomrule[1.1pt]
% \end{tabular}
% \end{table*}

\vspace{-5pt}
\subsection{Main Results}

\textbf{Overall comparison.}
Tables~\ref{tab:finance_prediction}--\ref{tab:weather_reasoning} summarize the main results under the unified {17-task / 30-metric} evaluation definition used throughout this paper. Under this setting, TimeClaw achieves the best overall result on {27 of the 30 reported metrics}. Broken down by task type, TimeClaw is best on {9/10 core numeric prediction and forecasting metrics}, {6/6 indicator metrics}, and {12/14 reasoning metrics}.

\textbf{Numeric prediction.}
We first focus on the core numeric tasks, namely finance prediction and weather forecasting. On finance prediction (\tabref{tab:finance_prediction}), TimeClaw improves over GPT-5.4 on all six reported metrics; the only exception is 30-day MACD MSE, where Llama-3 remains the strongest. On weather forecasting (\tabref{tab:weather_prediction}), TimeClaw achieves the best result on both forecast horizons: the 14-day/7-day weather forecast MSE drops from 20.984/14.197 to 13.525/10.719. This pattern supports our main motivation: when multiple executions are task-valid but quantitatively unequal, explicit metric-based comparison during exploratory learning can produce better downstream execution strategies.

\textbf{Indicator prediction.}
% We report weather-indicator prediction separately because it is numerically evaluated but somewhat orthogonal to the core forecast rows and can be moved to the appendix without changing the main story. 
% On this block, 
As shown in Table~\ref{tab:weather_prediction}, TimeClaw achieves the best result on all six reported metrics, covering max, min, and diff under both MSE and MAE. The gains are especially notable on the short setting, where TimeClaw improves all six cells over GPT-5.4.

\begin{wraptable}{r}{0.56\textwidth}
\vspace{-14pt}
\centering
\scriptsize
\setlength{\tabcolsep}{3.5pt}
\renewcommand{\arraystretch}{1.05}
\caption{Weather reasoning results.}
\label{tab:weather_reasoning}
\resizebox{\linewidth}{!}{
\begin{tabular}{lcccc}
\toprule[1.1pt]
Model & Past Trend & Future Trend & MCQA-S & MCQA-L \\
\cmidrule(lr){2-2}\cmidrule(lr){3-3}\cmidrule(lr){4-4}\cmidrule(lr){5-5}
 & Acc. & Acc. & Acc. & Acc. \\
\midrule[0.8pt]
GPT-4o    & 0.664 & 0.435 & 0.417 & 0.448 \\
Gemini    & 0.600 & 0.518 & 0.434 & 0.540 \\
DeepSeek  & 0.568 & 0.252 & 0.467 & 0.573 \\
Qwen-3    & 0.375 & 0.447 & 0.578 & 0.562 \\
Llama-3   & 0.473 & 0.197 & \underline{0.605} & \underline{0.626} \\
GPT-5.4   & \underline{0.980} & \underline{0.521} & 0.567 & 0.580 \\
TimeClaw  & \textbf{0.994} & \textbf{0.525} & \textbf{0.616} & \textbf{0.664} \\
\bottomrule[1.1pt]
\end{tabular}
}
\vspace{-10pt}
\end{wraptable}
\textbf{Reasoning tasks.}
On reasoning-oriented tasks, TimeClaw is best on 12/14 reported metrics. In finance reasoning (\tabref{tab:finance_all}), it substantially improves 7-day/30-day trend classification, both granularities of short-horizon correlation reasoning, and finance MCQA short; finance MCQA long remains very close to GPT-5.4. In weather reasoning (\tabref{tab:weather_reasoning}), it achieves the best results on past-trend prediction, future-trend prediction, and both MCQA settings.

The main remaining weakness is long-horizon finance correlation, where the strongest baseline still outperforms TimeClaw on 5-way/3-way label spaces. 

Overall, the results show that hierarchical distilled experience improves numeric prediction and multi-step decision procedures requiring temporal evidence, textual context, and tool outputs.
%Overall, the main results show that hierarchical distilled experience helps not only numeric prediction, but also multi-step decision procedures that require aligning temporal evidence, textual context, and tool outputs.
\vspace{-4pt}
\subsection{Ablation Study}

\begin{wraptable}{r}{0.58\textwidth}
\vspace{-16pt}
\centering
\small
\setlength{\tabcolsep}{4pt}
\caption{Component analysis on four tasks. \textit{w/o Exp.} disables hierarchical distilled experience injection.}
\label{tab:case1_ablation_4tasks}
\resizebox{\linewidth}{!}{
\begin{tabular}{llccc}
\toprule
Task & Metric & GPT-5.4 & w/o Exp. & TimeClaw \\
\midrule
Weather Forecast Short & MSE $\downarrow$ & 14.197 & \underline{13.058} & \textbf{10.719} \\
Finance Correlation Short & Acc.-3 $\uparrow$ & 0.448 & \underline{0.564} & \textbf{0.580} \\
Weather MCQA Long & Acc. $\uparrow$ & 0.580 & \underline{0.596} & \textbf{0.664} \\
Finance Trend Short & Acc. $\uparrow$ & 0.569 & \underline{0.663} & \textbf{0.696} \\
\bottomrule
\end{tabular}
}
\vspace{-10pt}
\end{wraptable}
Table~\ref{tab:case1_ablation_4tasks} shows a consistent ordering: GPT-5.4 $<$ \textit{noexp} $<$ \textsc{TimeClaw}. 
This pattern separates two confounded effects. 
First, moving from GPT-5.4 to \textit{noexp} shows the value of agent runtime and tool access. 
Second, moving from \textit{noexp} to \textsc{TimeClaw} isolates learned hierarchical distilled experience. 
These results support the claim that TimeClaw's advantage comes not only from tool exposure, but from learning reusable execution strategies and reinjecting them at inference time.
% \label{sec:case-study}

\begin{wraptable}{r}{0.58\textwidth}
\vspace{-14pt}
\centering
\small
\setlength{\tabcolsep}{4pt}
\caption{Paired comparison on the same samples.}
\label{tab:case_study_skills_only_weather}
\resizebox{\linewidth}{!}{
\begin{tabular}{lcccc}
\toprule
Task & Metric & GPT-5.4 & w/o Mem. & TimeClaw \\
\midrule
weather\_forecast\_long & RMSE $\downarrow$ & 4.581 & 3.973 & 3.753 \\
weather\_trend\_future\_short & Accuracy $\uparrow$ & 0.423 & 0.437 & 0.479 \\
\bottomrule
\end{tabular}
}
\vspace{-10pt}
\end{wraptable}
To isolate hierarchical distilled experience, we compare the full model with a \texttt{w/o Mem.} ablation that removes higher-level distilled layers while retaining tool-use routines. 
Table~\ref{tab:case_study_skills_only_weather} shows that TimeClaw outperforms \texttt{w/o Mem.} on two weather tasks. 
This suggests that TimeClaw's gains come not only from retaining experience, but from organizing it into layers that improve retrieval and guidance.
\vspace{-11pt}
\subsection{Tool-use Learning Analysis}
Figure~\ref{fig:case-study-macd-biocryst} shows how exploration changes the agent's strategy on a finance forecasting task. 
The unexplored agent maps bearish news and a crash to a negative MACD forecast. 
In contrast, the explored agent uses hierarchical distilled experience to decompose the task: it treats news as bias, detects an abnormal crash block, forecasts stabilization, and derives the MACD trajectory. 
This case shows that TimeClaw improves prediction quality and multi-step execution.

% \begin{table}[t]
% \centering
% \small
% \begin{tabular}{lccccc}
% \toprule
% Task & Metric & w/o xxx & Full Hierarchy  \\
% \midrule
% weather\_forecast\_long & RMSE $\downarrow$ & 3.973 & 3.753 \\
% weather\_trend\_future\_short  & Accuracy $\uparrow$ & 0.437 & 0.479\\
% \bottomrule
% \end{tabular}
% \caption{Paired comparison on the same samples. Skills-only underperforms the full hierarchical experience design on both tasks, with a larger gap on long-horizon weather forecasting.}
% \label{tab:case_study_skills_only_weather}
% \end{table}

\clearpage
\begin{figure}[h]
\centering
\resizebox{0.92\textwidth}{!}{%
\begin{minipage}{\textwidth}
\begin{minipage}[t]{1.00\textwidth}
\begin{tcolorbox}[
  colback=gray!10,
  colframe=gray!55,
  boxrule=0.4pt,
  arc=2pt,
  left=3pt, right=3pt, top=2pt, bottom=2pt,
  fontupper=\scriptsize\raggedright
]
\textbf{Task}: Predict the next 56 MACD values from a 5-minute stock series and a news article.\par\vspace{2pt}

\textbf{Input Focus}: Price tail:
\texttt{17.94, 17.82, 13.97, 12.95, 12.69, 12.52, 12.26, 12.06}.\par
News:
BioCryst pauses three BCX9930 studies after elevated creatinine / possible kidney-safety concerns.\par\vspace{2pt}

\textbf{True Future Pattern}: The ground-truth MACD does \emph{not} continue the crash monotonically. It stays mildly negative and gradually recovers toward near-zero
(tail around \texttt{-0.09}).
\end{tcolorbox}
\end{minipage}

\vspace{2pt}

\begin{minipage}[t]{0.30\textwidth}
\begin{tcolorbox}[
  colback=red!3,
  colframe=red!45!black,
  boxrule=0.4pt,
  arc=2pt,
  left=3pt, right=3pt, top=2pt, bottom=2pt,
  fontupper=\scriptsize\raggedright
]
\textbf{TimeClaw Unexplored}\par\vspace{2pt}

\textbf{Shortcut}\par
Negative news + sharp crash $\Rightarrow$ MACD remain strongly negative.\par\vspace{2pt}

\textbf{Reasoning}\par
It directly maps the bearish headline and recent drop into a persistently over-negative momentum forecast, without checking whether the crash is already an exhausted shock.\par\vspace{2pt}

\textbf{Tools}\par
\texttt{llm\_reasoning\_indicator}\par\vspace{2pt}

\textbf{Prediction}\par
\texttt{[-1.024, ..., -0.351]}\par\vspace{2pt}

It extends crash monotonically, while the true future MACD rebounds toward near-zero.\\
\texttt{macd\_mse = 0.486}
\end{tcolorbox}
\end{minipage}
\hfill
\begin{minipage}[t]{0.69\textwidth}
\begin{tcolorbox}[
  colback=blue!3,
  colframe=blue!45!black,
  boxrule=0.4pt,
  arc=2pt,
  left=3pt, right=3pt, top=2pt, bottom=2pt,
  fontupper=\scriptsize\raggedright
]
\textbf{TimeClaw Explored}\par\vspace{2pt}

\textbf{Learned Rule}\par
Use news only for \emph{directional bias}; let \emph{post-shock price structure} determine the MACD path.
After an event-driven crash block, momentum stabilizes or fades.\par\vspace{2pt}

\textbf{Reasoning Steps}\par
1. \texttt{sentiment} confirms that the article is a strong bearish catalyst.\par
2. \texttt{detect\_anomaly} finds that the last 21 points form a single abnormal crash block rather than an ordinary smooth trend.\par
3. \texttt{chronos2\_forecast} predicts stabilization around \texttt{12.25--12.56}, indicating that the immediate shock is being absorbed instead of continuing as free fall.\par
4. \texttt{llm\_reasoning\_indicator} converts this stabilized post-shock path into a MACD trajectory that starts negative but recovers toward near-zero.\par\vspace{2pt}

\textbf{Tools}\par
\texttt{sentiment -> detect\_anomaly ->}
\texttt{chronos2\_forecast -> llm\_reasoning\_ind.}\par\vspace{2pt}

\textbf{Prediction}\par
\texttt{[-0.849, -0.785, ..., 0.024]}\par\vspace{2pt}

Prediction matches the true negative-to-recovery shape, reducing error by \textbf{5.5$\times$}.\par
\texttt{macd\_mse = 0.088}
\end{tcolorbox}
\end{minipage}
\end{minipage}
}
\vspace{-4pt}
\caption{{Case study of post-shock MACD forecasting.}
The figure compares TimeClaw Unexplored and TimeClaw Explored on a finance task involving a stock-price crash and related news.}
\label{fig:case-study-macd-biocryst}
\vspace{-8pt}
\end{figure}

% \begin{figure}[t]
%     \centering

%     % Top panel
%     \begin{minipage}[t]{0.98\textwidth}
%         \centering
%         \includegraphics[width=\textwidth]{Figures/tool_dropout1.pdf}
        
%         \vspace{0.25em}
%         {\small (a) \textbf{Tool-Dropout Mechanism}.}
%     \end{minipage}

%     \vspace{0.6em}

%     % Bottom-left panel
%     \begin{minipage}[t]{0.485\textwidth}
%         \centering
%         \includegraphics[width=\textwidth]{Figures/tool_dropout2.pdf}
        
%         \vspace{0.25em}
%         {\small (b) \textbf{Tool Concentration Over Exploration}.}
%     \end{minipage}
%     \hfill
%     % Bottom-right panel
%     \begin{minipage}[t]{0.485\textwidth}
%         \centering
%         \includegraphics[width=\textwidth]{Figures/tool_dropout3.pdf}
        
%         \vspace{0.25em}
%         {\small (c) \textbf{Early Tool Coverage}.}
%     \end{minipage}

%     \vspace{0.5em}
%     \caption{\textbf{Tool dropout improves exploration and reduces concentration during exploration.}
%     The top panel illustrates the tool-dropout mechanism, where frequently used tools are temporarily masked according to their historical usage counts. The bottom panels show that tool dropout increases early tool coverage and reduces the long-run concentration of tool calls on a small set of dominant tools.}
%     \label{fig:case2_all}
% \end{figure}

\begin{figure}[h]
    \centering
    \includegraphics[width=0.8\textwidth]{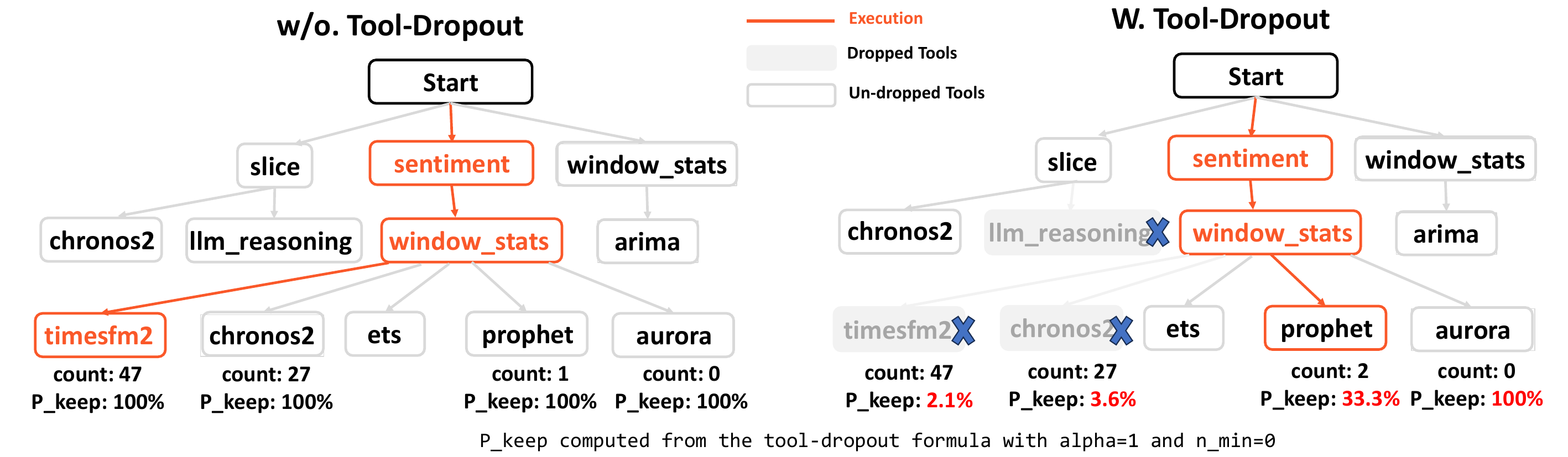}
    \caption{\textbf{Tool-dropout mechanism.}
    Frequently used tools are temporarily masked according to their historical usage counts to encourage broader exploration over candidate tools.}
    \label{fig:tool_dropout_mechanism}
\end{figure}

\subsection{Tool Dropout Analysis}
\begin{wrapfigure}{r}{0.43\textwidth}
    \vspace{-36pt}
    \centering

    \begin{minipage}{\linewidth}
        \centering
        \includegraphics[width=0.8\linewidth]{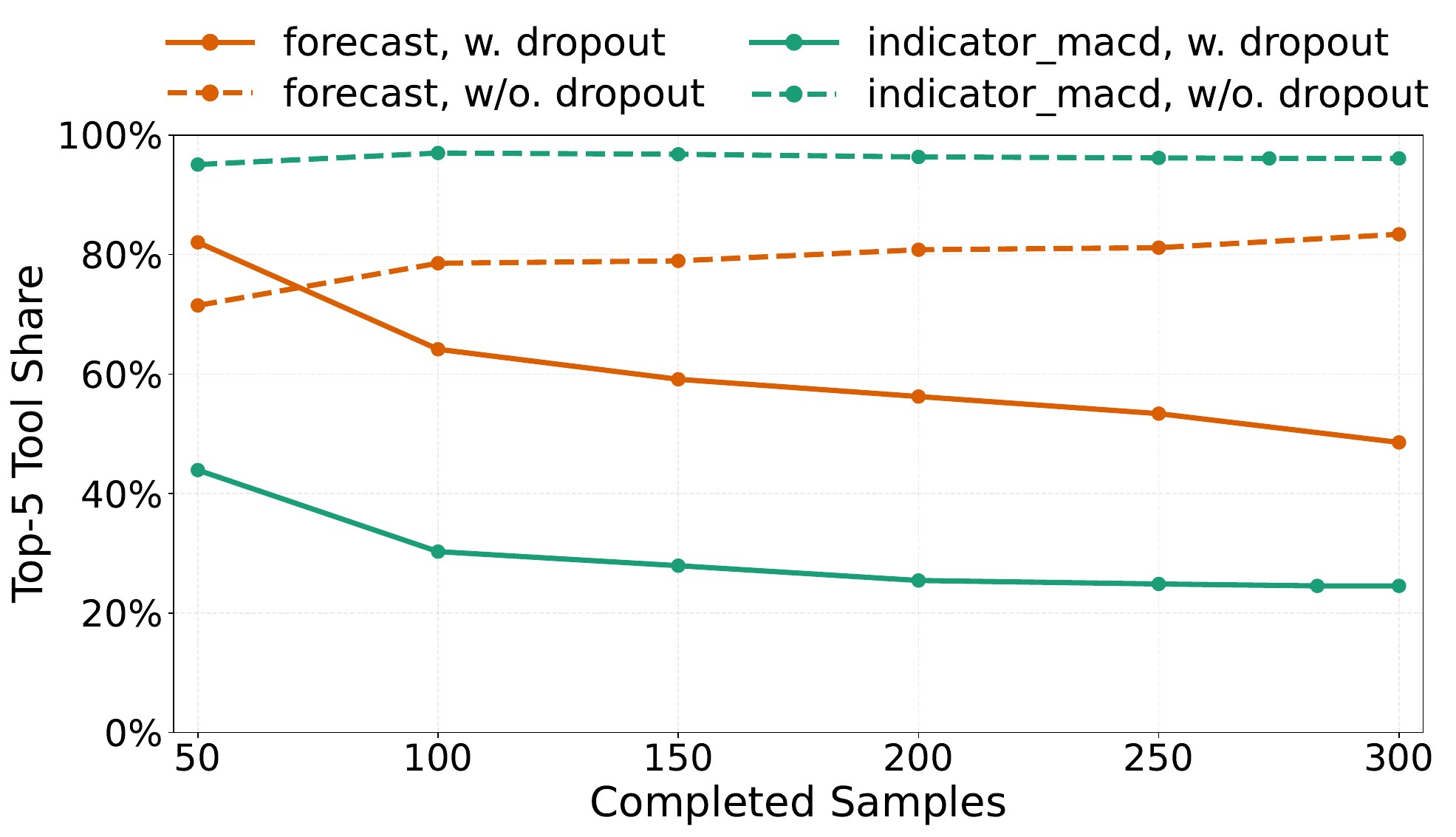}
        
        % \vspace{0.1em}
        % {\scriptsize (b) \textbf{Tool Concentration Over Exploration}.}
    \end{minipage}

    \vspace{0.5em}

    \begin{minipage}{\linewidth}
        \centering
        \includegraphics[width=0.8\linewidth]{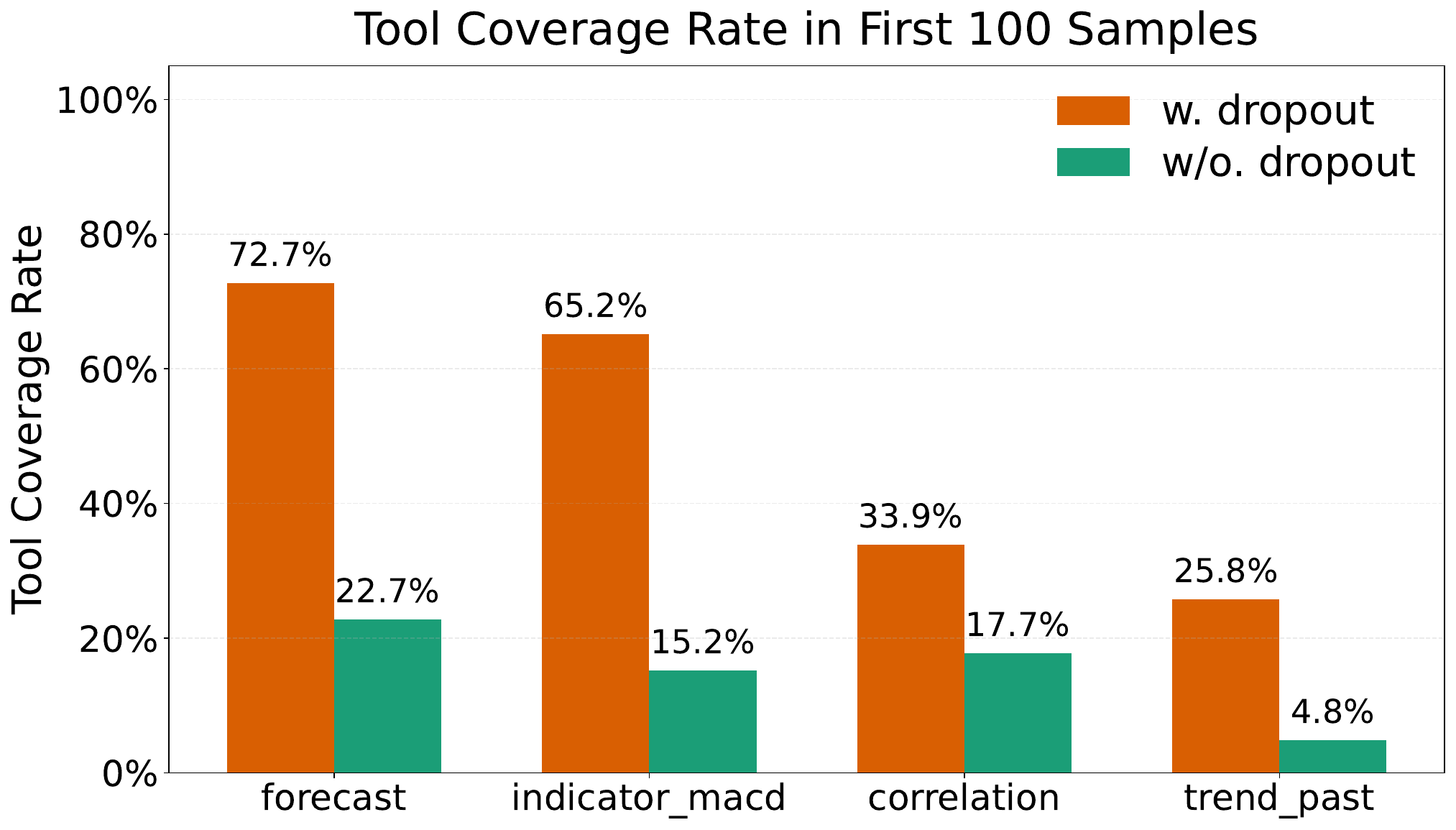}
        
        % \vspace{0.1em}
        % {\scriptsize (c) \textbf{Early Tool Coverage}.}
    \end{minipage}

    \vspace{0.3em}
\caption{Tool exploration analysis.}
    \label{fig:tool_dropout_bc}
    \vspace{-10pt}
\end{wrapfigure}
We evaluate whether tool dropout mitigates tool-prior collapse during exploratory execution learning on four tasks: \texttt{forecast}, \texttt{indicator\_macd}, \texttt{correlation}, and \texttt{trend\_past}. 
We use two behavioral metrics: \emph{tool coverage rate}, measuring how many visible tools are invoked within an exploration prefix, and \emph{Top-5 tool share}, measuring concentration on the five most-used tools. 
Figure~\ref{fig:tool_dropout_mechanism} illustrates how dropout masks historically dominant tools to encourage alternative tool chains. 
Figure~\ref{fig:tool_dropout_bc} shows that this intervention reduces tool-use concentration and increases early coverage, indicating broader exploration.
Overall, tool dropout reduces reliance on dominant tools and increases early tool coverage, indicating that it preserves exploration diversity and mitigates premature tool-prior collapse.

\section{Conclusion}

We presented {TimeClaw}, a time-series agent framework that learns from \emph{exploratory execution}. 
It compares task-valid executions under verifiable metrics, distills evidence into hierarchical external experience, and reuses it at inference time without online adaptation or parameter updates. 
Across the MTBench setting, TimeClaw improves both numeric prediction and reasoning, showing that verifiable time-series agents should learn to compare, organize, and reuse exploratory experience.
%More broadly, TimeClaw provides a practical route to scientific-domain agents that improve through accumulated execution strategies rather than continual weight updates.

% We presented \textbf{TimeClaw}, a time-series agent framework that learns from \emph{exploratory execution}. Instead of treating each instance as an isolated inference problem, TimeClaw compares candidate executions under task metrics, distills the resulting evidence into hierarchical external experience, and reuses that experience at inference time without online adaptation or parameter updates.

% Our results show that, for verifiable time-series tasks, the key challenge is not only execution-time capability, but how exploratory experience is compared, organized, and reused. Across the unified 17-task / 30-metric MTBench setting used in this paper, TimeClaw achieves broad gains on both numeric prediction and reasoning tasks. More broadly, TimeClaw suggests a practical direction for scientific-domain agents: improving not by continual weight updates, but by explicitly accumulating and reusing better execution strategies learned from exploration.

\newpage

\bibliographystyle{IEEEtran}
\bibliography{reference-truth}
\newpage
% \input{appendix.tex}

%%%%%%%%%%%%%%%%%%%%%%%%%%%%%%%%%%%%%%%%%%%%%%%%%%%%%%%%%%%%

\appendix

\section{Implementation Details}
\label{app:implementation}

This section records implementation-specific details that are not central to the main text. Exploratory learning and downstream inference share the same core execution pipeline and tool stack, but differ in permissions: exploration can access supervision tools and update external artifacts such as notes, memory, and skills, whereas inference reuses the distilled artifacts in read-only form. Our results focus on the procedural shifts in agent reasoning, which we validate through detailed component analysis (noexp vs. TimeClaw) and qualitative trace evidence that confirms the reliability of the learned strategies. We observe that the performance gains are tied to these structural improvements in the execution policy rather than random variance. In our experiments, language-model calls are served through an OpenAI-compatible external API interface, while the local time-series components, including the forecasting and indicator backbones used by the agent tools, are deployed on two NVIDIA RTX 3090 GPUs. We explore 300 samples per task, and a full exploration run takes about one day in the current setup. In the current setup, a full exploration run takes a substantial number of tokens due to the generation of multiple candidate paths for each
sample. 
% Our corpus and code is available at \url{https://anonymous.4open.science/r/TimeClaw_NIPS2026-631B}.

\section{Dataset Details}
\label{app:data}

This section provides appendix-only details on corpus construction, with emphasis on source independence and task coverage.

\subsection{Learning Corpus}
\label{app:data:learning}

The learning corpus is constructed to be MTBench-compatible rather than benchmark-identical. Concretely, we align the corpus to the MTBench task contracts, task families, horizon structure, label spaces, and input-output formats, so that exploratory execution is performed under the same problem interface that will later appear at evaluation time. At the same time, the underlying samples are generated from an independent data pipeline rather than copied from the official benchmark.

The independent construction follows the two domains used in the paper. On the weather side, we build aligned forecasting, trend, indicator, and MCQA-style instances from separate weather and event sources. On the finance side, we build aligned forecasting and reasoning instances from separately collected price and news data, and match the benchmark at the level of sequence-length schedules, answer schemas, and task-specific output structure. For label-based tasks, the generated pool is further rebalanced so that the marginal task distribution remains comparable to the benchmark setting.

To ensure that transfer is measured rather than memorization, the learning corpus is explicitly isolated from the benchmark evaluation samples. In particular, the construction enforces source-level disjointness checks, such as non-overlap of finance URLs and weather station identifiers, while preserving schema compatibility with the MTBench evaluation interface. This design allows TimeClaw to learn reusable execution experience on aligned but independent samples, and then test whether that experience improves performance on unseen benchmark instances.

\subsection{Evaluation Corpus}
\label{app:data:evaluation}

The evaluation corpus covers both weather and finance. In weather, we evaluate forecasting, future-trend classification, past-trend classification, indicator prediction, and multiple-choice question answering. In finance, we evaluate forecasting, trend classification, indicator prediction, multiple-choice question answering, and news--price correlation reasoning. Each domain includes short-horizon and long-horizon settings so that both near-term and extended-window reasoning can be tested under the same benchmark framework.

\subsection{Task Inventory}
\label{app:data:tasks}

For reference, the paper reports a fixed 17-task evaluation suite: four finance prediction tasks, three weather prediction tasks, six finance reasoning tasks, and four weather reasoning tasks. This paper-level inventory should be distinguished from internal execution tasks. Some tasks may involve multiple output fields or metric views, but are still counted as one paper task; for example, weather-indicator prediction is one task, even though it is reported with multiple numeric metrics.

For clarity, we summarize the task requirements in plain language below.
\begin{itemize}[leftmargin=1.5em]
    \item \textbf{Finance forecast (7D / 30D).} The agent receives a historical stock-price window together with related text context and must predict the future price sequence over the target horizon. In simple terms, the question is: given the recent market history, what will the next period of the price curve look like?
    \item \textbf{Finance MACD prediction (7D / 30D).} The agent receives the same kind of stock-history input, but the target is not the raw future price curve. Instead, it must predict the future MACD indicator trajectory. In simple terms, the task asks whether the agent can reason about a derived technical indicator rather than only the original series.
    \item \textbf{Weather forecast (7D / 14D).} The agent observes past weather measurements and must predict the future temperature sequence for the required horizon. In simple terms, the task is to continue the weather curve into the future.
    \item \textbf{Weather-indicator prediction.} The agent observes the historical weather window and predicts summary values of the future window, namely the future maximum, minimum, and temperature difference. In simple terms, the agent does not need to output the full future curve, but must summarize the important future extremes correctly.
    \item \textbf{Finance trend classification (7D / 30D).} Instead of predicting exact future values, the agent must decide which discrete trend label best describes the future stock movement. In simple terms, the task asks whether the future movement is strongly down, mildly down, neutral, mildly up, or strongly up.
    \item \textbf{Finance correlation reasoning (7D / 30D).} The agent is given market data together with finance news and must judge how the news and the future price movement are related. In simple terms, it must decide whether the relation is strongly positive, moderately positive, neutral, moderately negative, or strongly negative.
    \item \textbf{Finance MCQA (7D / 30D).} The agent is given a finance question with four candidate answers, together with the associated time-series and text context, and must choose the correct option. In simple terms, this is a benchmark multiple-choice reasoning task grounded in financial time-series evidence.
    \item \textbf{Weather future-trend classification.} The agent receives a past weather window and must decide the overall direction of the coming weather period. In simple terms, the question is whether the future weather is generally going up, going down, or staying stable.
    \item \textbf{Weather past-trend classification.} The agent receives a past weather window and must classify the trend already present inside that historical window. In simple terms, instead of forecasting the future, it must summarize whether the recent history itself was increasing, decreasing, or stable.
    \item \textbf{Weather MCQA (short / long).} The agent is given a weather question with four answer choices and must select the correct one using the time-series context. In simple terms, this tests weather-grounded multiple-choice reasoning rather than direct numeric prediction.
\end{itemize}

\begin{table}[t]
\centering
\small
\setlength{\tabcolsep}{5pt}
\renewcommand{\arraystretch}{1.05}
\caption{Plain-language summary of the 17 paper tasks.}
\label{tab:appendix_task_plain}
\begin{tabular}{p{0.26\linewidth}p{0.34\linewidth}p{0.28\linewidth}}
\toprule
Task & Input & Required output \\
\midrule
Finance forecast (7D / 30D) & Historical stock-price window with related text context & Future price sequence \\
Finance MACD prediction (7D / 30D) & Historical stock-price window with related text context & Future MACD indicator sequence \\
Weather forecast (7D / 14D) & Historical weather window & Future temperature sequence \\
Weather indicator & Historical weather window & Future max / min / diff summary values \\
Finance trend (7D / 30D) & Historical stock-price window with related text context & One future trend label \\
Finance correlation (7D / 30D) & Market series together with finance news & One correlation label \\
Finance MCQA (7D / 30D) & Finance question, options, and time-series/text context & One answer option \\
Weather future trend & Historical weather window & One future trend label \\
Weather past trend & Historical weather window & One historical trend label \\
Weather MCQA (short / long) & Weather question, options, and time-series context & One answer option \\
\bottomrule
\end{tabular}
\end{table}

\section{Evaluation Protocol}
\label{app:evaluation}

This section records the appendix-level scoring details behind the main-text results. All reported results follow the same 17-task MTBench evaluation suite, with official task prompts, output contracts, label spaces, and MTBench-compatible scoring conventions kept fixed.

\subsection{Metric Mapping}
\label{app:evaluation:metrics}

The paper reports 30 metrics induced by the 17-task suite. For numeric prediction tasks, we report the official task metrics associated with each task family: finance forecasting uses MAE and MAPE; finance MACD prediction uses MSE; weather forecasting uses MSE and MAE; and weather-indicator prediction uses max/min/diff under both MSE and MAE. For reasoning tasks, we report accuracy-based metrics. Finance trend and finance correlation are evaluated under both the original 5-way label space and the coarser 3-way grouping, while MCQA and weather trend tasks are reported with standard accuracy. When model outputs do not exactly match the target sequence length for numeric tasks, predictions are aligned to ground-truth length using the same MTBench-style interpolation rule before metric computation.

\subsection{Included and Filtered Samples}
\label{app:evaluation:filtering}

We distinguish raw recorded rows from rows that contribute to the final benchmark summary. A sample is excluded from summary aggregation if it is unscorable under the benchmark contract, for example, because the ground truth is invalid, the prediction cannot be parsed into the required output type, or execution terminates abnormally. For cumulative numeric summaries, we further apply the same official-style threshold filtering policy used in the MTBench-compatible evaluation path: rows may still be kept in the raw log for auditing, but extreme-error rows above the task-specific official threshold do not contribute to the final averaged metric. Therefore, the effective sample count used by a reported metric may be smaller than the number of raw inference rows, and this behavior reflects the benchmark's official summary policy rather than an additional TimeClaw-specific reporting rule.

\section{Details of Tools}
\label{app:tools}

TimeClaw exposes a relatively large tool library, so we organize the tools by functional role rather than enumerate every tool in a flat list. At a high level, the tool library contains forecasting tools, series-analysis tools, text-analysis tools, and exploration-only supervision tools. Different tasks activate different subsets of these tools, but the grouping below captures the main functional structure used by the agent.

\subsection{Forecasting and Indicator Tools}
\label{app:tools:forecast}

The first group contains forecasting-oriented tools that directly produce future trajectories or indicator predictions. These tools are most important for numeric prediction tasks such as stock forecasting, weather forecasting, and MACD prediction. Representative examples include \texttt{chronos\_forecast}, \texttt{moirai\_forecast}, \texttt{timesfm\_forecast}, \texttt{aurora\_forecast}, \texttt{timer\_forecast}, \texttt{prophet}, \texttt{ets}, and \texttt{arima}. In simple terms, these tools answer questions such as ``what does the future curve look like?'' or ``what will the future technical indicator trajectory be?'' They are often used together with lightweight preprocessing and evidence tools rather than in isolation.

\subsection{Reasoning and Analysis Tools}
\label{app:tools:reasoning}

The second group contains analysis tools that help the agent interpret observed data before producing a final answer. 
Profiling tools summarize statistical properties, detection tools identify temporal patterns, and relation-oriented tools measure dependencies among signals. 
Together, they help the agent assess trends, co-movement, abrupt changes, and whether candidate explanations are supported by the data. 
For text-grounded tasks, TimeClaw further uses text-analysis tools such as \texttt{summarize}, \texttt{keyword\_extract}, \texttt{sentiment}, \texttt{ner}, \texttt{text\_similarity}, and \texttt{temporal\_align\_text} to connect news or question text with time-series evidence, especially in finance correlation and MCQA tasks.

% The second group contains analysis tools that help the agent understand the structure of the observed data before deciding which final answer to produce. We use profiling tools such as \texttt{basic\_stats}, \texttt{autocorrelation}, \texttt{stationarity}, \texttt{hurst}, and \texttt{spectral\_density} to summarize the statistical properties of a time series. We use detection tools such as \texttt{trend}, \texttt{seasonality}, \texttt{changepoint}, \texttt{anomaly}, and \texttt{regime} to identify visible temporal patterns. For relation-oriented reasoning, we use tools such as \texttt{correlation}, \texttt{lagged\_corr}, \texttt{granger}, \texttt{cointegration}, \texttt{mutual\_info}, and \texttt{transfer\_entropy}. In simple terms, these tools help the agent decide whether a series is rising or falling, whether two signals move together, whether a sudden change has occurred, or whether a candidate explanation is supported by the observed data.

% For text-grounded tasks, TimeClaw uses text-analysis tools like \texttt{read\_raw\_text}, \texttt{summarize}, \texttt{keyword\_extract}, \texttt{sentiment}, \texttt{ner}, \texttt{text\_similarity}, and \texttt{temporal\_align\_text}. These tools are especially important for finance correlation and MCQA tasks, where the agent must connect news content or question text to the underlying time-series evidence.

\subsection{Exploration-Only Supervision and Orchestration Tools}
\label{app:tools:exploration_only}

The final group contains tools that are available only during exploratory learning and are not exposed at test time. The most important examples are \texttt{evaluate\_against\_gt} and its batch form, which compare a candidate prediction against ground truth and return task metrics for learning-time analysis. These tools are used to support candidate comparison and experience distillation, not to solve benchmark tasks directly. In addition, exploratory learning may expose branch-generation or orchestration capabilities that help produce multiple candidate paths during exploration. These exploration-only tools are removed at inference time, so the deployed agent solves benchmark tasks using only task-facing evidence tools and previously distilled experience.

\begin{tcolorbox}[
  colback=gray!10,
  colframe=gray!55,
  boxrule=0.5pt,
  arc=3pt,
  left=4pt, right=4pt, top=3pt, bottom=3pt,
  fontupper=\scriptsize\raggedright
]
\begin{verbatim}
## Objective
### Task Prompt
{task_prompt}
{task_boundary_section}
{execution_context_section}
{control_context_section}
{output_contract_section}

## Observation
### Sample Fingerprint
{fingerprint}
{profiling_section}
{support_section}

## Decision
{decision_section}

### Available Tools
{tool_names}
\end{verbatim}
\end{tcolorbox}

\subsection{Exploration-Time Context}
\label{app:prompt:explorer}

\begin{tcolorbox}[
  colback=blue!3,
  colframe=blue!45!black,
  boxrule=0.5pt,
  arc=3pt,
  left=4pt, right=4pt, top=3pt, bottom=3pt,
  fontupper=\scriptsize\raggedright
]
\textbf{Exploration-Time Prompt Fragment}\par\vspace{4pt}

\begin{verbatim}
## Spawn Guidance
- If you think this step would benefit from exploring multiple follow-up
  branches for comparison, prefer using spawn_subagent.
- In exploration mode, the first spawn round should create at least 2
  distinct candidate branches.

## Evaluate Guidance
- evaluate_* is for post-hoc validation and retrospective review,
  not for solving the task itself.
- Before finishing this learning run, evaluate at least one task-valid
  candidate.
- If multiple reasonable candidates exist, compare at least 2 candidate
  paths before finishing unless tool availability makes that impossible.
- If you finish the learning run itself, answer_type must be
  "learning_summary".

## Control Context
- state = need_evidence
- required_final_type = learning_summary
- subagent slots / branch hints if branch exploration is enabled

## Support
- official task description
- sample fingerprint and profiling block
- optional raw-text access note: call read_raw_text if needed
\end{verbatim}
\end{tcolorbox}

\subsection{Inference-Time Context}
\label{app:prompt:infer}

\begin{tcolorbox}[
  colback=green!3,
  colframe=green!45!black,
  boxrule=0.5pt,
  arc=3pt,
  left=4pt, right=4pt, top=3pt, bottom=3pt,
  fontupper=\small\raggedright
]
\textbf{Inference-Time Prompt Fragment}\par\vspace{4pt}

\begin{verbatim}
## Control Context
- state = need_evidence
- required_final_type = {task_type}

## Support
- sample-level selected experience packet
- selected skills
- selected decision skills
- selected memory rules
- focused tool notes for selected tools only

### Available Tools
- task-facing tools only
- no spawn_subagent
- no evaluate_against_gt / evaluate_batch_against_gt

## Completion
- ordinary benchmark answer contract
- no learning_summary
\end{verbatim}
\end{tcolorbox}

\subsection{Representative Prompt Fragments}
\label{app:prompt:fragments}

\begin{tcolorbox}[
  colback=orange!3,
  colframe=orange!50!black,
  boxrule=0.5pt,
  arc=3pt,
  left=4pt, right=4pt, top=3pt, bottom=3pt,
  fontupper=\small\raggedright
]
\textbf{Representative Runtime Content}\par\vspace{4pt}

\begin{verbatim}
Task Prompt:
official MTBench system/user prompt pair merged into one description block

Sample Fingerprint:
compact numeric fingerprint of the current sample

Profiling Section:
basic_stats, autocorrelation, stationarity_test,
detect_trend, detect_anomaly (when available)

Support Section:
reusable guidance + focused tool notes; in inference this is a filtered
sample-level packet rather than the full stored memory

Sub-Agent Variant:
same prompt frame, but branch-local goal and slot-local tool hint are added
so different branches explore different candidate paths
\end{verbatim}
\end{tcolorbox}

\section{External Learning and Distillation}
\label{app:distill}

This section records the implementation details that are useful for reproducing the current exploration pipeline.

\subsection{From Exploratory Execution to Notes}
\label{app:distill:notes}

One exploratory exploration episode is first committed into the append-only \texttt{Notes} layer. In the current implementation, the pipeline writes a note only after the run has produced usable evaluation evidence. Concretely, the execution trace must contain at least one \texttt{evaluate\_against\_gt} or batch-evaluation result from which the system can extract task metrics.

Before the note is written, the pipeline normalizes the final \texttt{learning\_summary} against the traced candidate results so that the stored winner, tool chain, and summary text remain aligned with the actually evaluated path. It then constructs a structured single-sample learning analysis from the candidate traces, selected metrics, and task context, and renders that analysis into the note text.

The stored note includes the task key, task prompt, compact numeric time-series block, selected text signals, winner tools, recorded prediction, recorded metrics, sample fingerprint, pipeline summary, candidate traces, and source metadata. Notes are written in append-only form and sharded by exact task under \texttt{notes/<task>.md}.

\subsection{From Notes to Memory, Tool Notes, and Skills}
\label{app:distill:memory}

After a note is committed, a task-local distillation worker is triggered. The first stage is \texttt{Notes $\rightarrow$ Memory}. In the current implementation, this stage is triggered every 10 newly accumulated notes for the same exact task, although explicit finalization can also flush shorter tails. During this step, the system gathers the pending note slice, resolves note-level conflicts, computes summary statistics, and asks the distillation model to update a bounded set of structured memory rules. The resulting rules are stored in the task-local \texttt{Memory} layer, with a hard cap of 30 rules per task.

The next stages are derived from the latest full \texttt{Memory} view rather than directly from raw notes. \texttt{Memory $\rightarrow$ Tool Notes} rebuilds per-tool boundary cards. \texttt{Memory $\rightarrow$ Skills} writes compact operational SOP-style guidance. \texttt{Memory $\rightarrow$ Skills Decision} writes compact decision-oriented guidance. These three stages are refreshed when the task-local memory fingerprint changes, so they are rebuilt only when the upstream memory content has actually changed.

An implementation detail that matters for reproducibility is that writing an artifact and injecting an artifact are two different steps. The exploration pipeline refreshes \texttt{Memory}, \texttt{Tool Notes}, \texttt{Skills}, and \texttt{Skills Decision} as persistent stores; later inference may retrieve them, but still performs an additional sample-level selection step before prompt injection.

\subsection{Task-Aware Tool Dropout}
\label{app:distill:dropout}

Task-aware tool dropout acts only during exploratory learning. After each exploration run, the system parses the execution trace, collects the actually used substantive tools, and updates per-task usage counts. When a later exploration run spawns sub-agents for the same task, these counts are used to precompute slot-local visible tool subsets. Frequently used tools are more likely to be suppressed, while underused tools remain visible.

The resulting slot specification is then passed into the exploratory prompt as branch-local tool availability.

\section{Learning and Inference Algorithms}
\label{Algorithms}
\RestyleAlgo{ruled}
\begin{algorithm2e}[t]
\caption{TimeClaw Exploratory Execution Learning Loop}
\label{alg:learn}
\KwIn{Exploration corpus $\mathcal{D}=\{x_1,\dots,x_N\}$; initial reusable experience state $\mathcal{A}_s^{0}$; task-dependent evaluation rule for $\tau$}
\KwOut{Learned reusable experience states $\{\mathcal{A}_s\}$}
\ForEach{Exploration instance $x=(z,c,\tau,s)\in\mathcal{D}$}{
    Retrieve current reusable experience state $\mathcal{A}_s$\;
    Run executor in exploration mode with exploration tools\;
    Generate $K\!\ge\!2$ candidate executions $\Pi(x)=\{\pi_1,\dots,\pi_K\}$\;
    Filter to task-valid subset: $\Pi_{\mathrm{valid}}(x)\!\leftarrow\!\{\pi_k : \hat{y}_k \text{ satisfies contract of } \tau\}$\;
    \uIf{$|\Pi_{\mathrm{valid}}(x)|\ge 2$}{
        Compute task-dependent execution quality $q(\pi_k;x)$\;
        Select $\pi^\star(x)=\argmax_{\pi_k\in\Pi_{\mathrm{valid}}(x)} q(\pi_k;x)$\;
    }
    \uElseIf{$|\Pi_{\mathrm{valid}}(x)|= 1$}{
        Record single-execution evidence (no comparative signal)\;
    }
    \Else{
        Record failure evidence; \textbf{continue}\;
    }
    Construct $n_x=\mathrm{Summarize}(x,\Pi(x),\{q(\pi_k;x)\},\pi^\star(x))$\;
    Clean to $e_x=\mathrm{Clean}(n_x)$\;
    Update reusable experience state: $\mathcal{A}_s \leftarrow \mathrm{Distill}(\mathcal{A}_s, e_x)$\;
}
\Return{$\{\mathcal{A}_s\}$}
\end{algorithm2e}

\begin{algorithm2e}[t]
\caption{TimeClaw Inference-Time Reuse}
\label{alg:infer}
\KwIn{New instance $x'=(z',c',\tau',s')$; learned reusable experience states $\{\mathcal{A}_s\}$}
\KwOut{Task prediction $\hat{y}$}
\Begin{
Retrieve reusable experience for scope $s'$\;
Filter to prompt-eligible reusable experience using scope, applicability, and injection status\;
Assemble prompt with task description, output contract, and sample fingerprint\;
Add profiling, \textsc{Soul}, scope-local \textsc{Memory}, tool notes, and \textsc{Skills}\;
Expose only runtime-compatible tools\;
Run the agent with retrieved experience and runtime tools\;
\Return{$\hat{y}$}\;
}
\end{algorithm2e}

\section{Qualitative Examples}
\label{app:qualitative}

This section provides artifact-grounded qualitative examples from both exploratory learning and downstream inference. We keep these examples intentionally concrete: every training case is reconstructed from a real training trace, and every inference case is reconstructed from a real row-level inference artifact. We do not add post-hoc interpretations that are absent from the stored artifacts. Instead, we show exactly what the system executed, what evidence each branch or tool produced, and what final record was written back into the trace or row artifact.

\subsection{Training Examples}

These training examples are reconstructed directly from real exploratory traces. We keep the presentation intentionally concrete: each case shows the task setup, the two realized branches, the recorded ground-truth comparison, and the final learning summary written back into the trace. To keep the figures readable, every displayed training case contains exactly two sub-agents. The purpose here is not to retell the task abstractly, but to let the reader inspect how TimeClaw compares alternative execution paths, identifies which path actually matches the task target, and records a reusable preference for future runs.

\begin{figure*}[t]
  \centering
  \begin{minipage}[t]{1.00\textwidth}
  \begin{tcolorbox}[
    colback=gray!10,
    colframe=gray!55,
    boxrule=0.5pt,
    arc=3pt,
    left=4pt, right=4pt, top=3pt, bottom=3pt,
    fontupper=\small\raggedright
  ]
  \textbf{Training Example 1}\par\vspace{4pt}
  
  \textbf{Task / Sample}\par
  \texttt{weather\_aligned\_long:USW00003017\_11.json:2:forecast}\par
  Task: predict the next 72 hourly temperature values from a 14-day hourly weather series and aligned severe-weather text.\par\vspace{4pt}
  
  \textbf{Task metadata and text excerpts used below}\par
  The trace records a weather long forecast task at station \texttt{USW00003017}. The text includes hail and thunderstorm-wind reports near the forecast boundary on \texttt{2011-07-13}, with heavy rain, strong wind, and storm damage across the Urban Corridor.\par
  \end{tcolorbox}
  \end{minipage}
  \vspace{4pt}
  \begin{minipage}[t]{1.00\textwidth}
  \begin{tcolorbox}[
    colback=blue!3,
    colframe=blue!45!black,
    boxrule=0.5pt,
    arc=3pt,
    left=4pt, right=4pt, top=3pt, bottom=3pt,
    fontupper=\small\raggedright
  ]
  \textbf{Main Agent}\par\vspace{4pt}
  
  \textbf{Spawn result}\par
  \texttt{spawn\_subagents: n\_tasks = 2}\par
  \texttt{sub-agent 0}: forecast the next 72 hourly temperatures with a \texttt{timesfm2\_forecast}-anchored path.\par
  \texttt{sub-agent 1}: forecast the next 72 hourly temperatures with a \texttt{chronos2\_forecast}-anchored path.\par
  The main trace later compared both task-valid 72-step forecasts against ground truth.\par
  \end{tcolorbox}
  \end{minipage}
  \vspace{4pt}
  \begin{minipage}[t]{0.48\textwidth}
  \begin{tcolorbox}[
    colback=red!3,
    colframe=red!45!black,
    boxrule=0.5pt,
    arc=3pt,
    left=4pt, right=4pt, top=3pt, bottom=3pt,
    fontupper=\small\raggedright
  ]
  \textbf{Sub-agent 0}\par
  \textbf{Tool chain}\par
  \texttt{timesfm2\_forecast}\par\vspace{4pt}
  
  \textbf{Sub-agent result}\par
  \texttt{answer = [26.578, 25.740, ..., 25.479]}\par\vspace{4pt}
  
  \textbf{Reasoning text}\par
  \small\raggedright
  \texttt{timesfm2\_forecast} was run on the original \texttt{336}-point hourly temperature series with horizon \texttt{72} and returned a full task-valid forecast array.\par
  
  The recorded explanation says this candidate preserved a realistic day-night cycle across all three forecast days and stayed in a plausible summer temperature range.\par
  
  Its first forecast day cooled into the high teens before reheating the next afternoon, matching the storm-adjacent cooling pattern discussed later in the learning summary.
  \end{tcolorbox}
  \end{minipage}
  \hfill
  \begin{minipage}[t]{0.50\textwidth}
  \begin{tcolorbox}[
    colback=green!3,
    colframe=green!45!black,
    boxrule=0.5pt,
    arc=3pt,
    left=4pt, right=4pt, top=3pt, bottom=3pt,
    fontupper=\small\raggedright
  ]
  \textbf{Sub-agent 1}\par
  \textbf{Tool chain}\par
  \texttt{chronos2\_forecast}\par\vspace{4pt}
  
  \textbf{Sub-agent result}\par
  \texttt{answer = [27.500, 26.250, ..., 25.875]}\par\vspace{4pt}
  
  \textbf{Reasoning text}\par
  \small\raggedright
  \texttt{chronos2\_forecast} was run on the same original \texttt{336}-point hourly series and also returned a full task-valid \texttt{72}-step forecast.\par
  
  The recorded explanation says this candidate preserved the diurnal cycle but with visibly quantized quarter-step values across much of the horizon.\par
  
  It also stayed warmer than the TimesFM path through the early post-storm cooling segment, which later became part of the failure analysis in the trace.
  \end{tcolorbox}
  \end{minipage}
  \vspace{4pt}
  \begin{minipage}[t]{1.00\textwidth}
  \begin{tcolorbox}[
    colback=purple!3,
    colframe=purple!45!black,
    boxrule=0.5pt,
    arc=3pt,
    left=4pt, right=4pt, top=3pt, bottom=3pt,
    fontupper=\small\raggedright
  ]
  \textbf{Ground truth Comparison and Final Learning Summary}\par\vspace{4pt}
  
  \textbf{Ground truth evaluation}\par
  \begin{quote}
  \small
  \texttt{sub-agent 0 / timesfm2\_forecast}: \texttt{MAE = 1.585}, \texttt{RMSE = 2.277}, \texttt{MAPE = 7.765}, \texttt{MSE = 5.183}\par
  \texttt{sub-agent 1 / chronos2\_forecast}: \texttt{MAE = 1.821}, \texttt{RMSE = 2.450}, \texttt{MAPE = 9.105}, \texttt{MSE = 6.001}
  \end{quote}
  
  \textbf{Final learning summary}\par
  \begin{quote}
  \small
  \texttt{insight}: Among the two evaluated \texttt{72}-step weather forecasts, the TimesFM branch performed best against ground truth, with lower error than the Chronos-2 branch on every reported metric. Both candidates captured the diurnal cycle, but the TimesFM forecast tracked the realized temperatures more closely overall.\par\vspace{4pt}
  
  \texttt{recommendation}: For similar hourly temperature forecasting with strong daily seasonality and storm-related context, prefer the \texttt{timesfm2\_forecast} path as the primary baseline and keep \texttt{chronos2\_forecast} as a secondary comparison model.
  \end{quote}
  \end{tcolorbox}
  \end{minipage}
  \caption{Execution sample on weather forecast task.}
  \end{figure*}
  
  \begin{figure*}[t]
  \centering
  \begin{minipage}[t]{1.00\textwidth}
  \begin{tcolorbox}[
    colback=gray!10,
    colframe=gray!55,
    boxrule=0.5pt,
    arc=3pt,
    left=4pt, right=4pt, top=3pt, bottom=3pt,
    fontupper=\small\raggedright
  ]
  \textbf{Training Example 2}\par\vspace{4pt}
  
  \textbf{Task / Sample}\par
  \texttt{finance\_aligned\_long:00006.parquet\_row:6:forecast}\par
  Task: predict the next 33 hourly stock prices from a 30-day hourly price series and an aligned finance article.\par\vspace{4pt}
  
  \textbf{Task metadata and text excerpts used below}\par
  The trace records a finance long forecast task on \texttt{AAP}. The article is mixed-to-neutral: it mentions strong prior earnings and raised guidance, but also says estimates flatlined and gives a \texttt{Zacks Rank \#3 (Hold)} outlook.\par
  \end{tcolorbox}
  \end{minipage}
  \vspace{4pt}
  \begin{minipage}[t]{1.00\textwidth}
  \begin{tcolorbox}[
    colback=blue!3,
    colframe=blue!45!black,
    boxrule=0.5pt,
    arc=3pt,
    left=4pt, right=4pt, top=3pt, bottom=3pt,
    fontupper=\small\raggedright
  ]
  \textbf{Main Agent}\par\vspace{4pt}
  
  \textbf{Spawn result}\par
  \texttt{spawn\_subagents: n\_tasks = 2}\par
  \texttt{sub-agent 0}: forecast the next 33 hourly prices with a \texttt{timesfm2\_forecast}-anchored path.\par
  \texttt{sub-agent 1}: forecast the next 33 hourly prices with a \texttt{chronos2\_forecast}-anchored path.\par
  The main trace later compared both task-valid forecast arrays against ground truth.\par
  \end{tcolorbox}
  \end{minipage}
  \vspace{4pt}
  \begin{minipage}[t]{0.50\textwidth}
  \begin{tcolorbox}[
    colback=red!3,
    colframe=red!45!black,
    boxrule=0.5pt,
    arc=3pt,
    left=4pt, right=4pt, top=3pt, bottom=3pt,
    fontupper=\small\raggedright
  ]
  \textbf{Sub-agent 0}\par
  \textbf{Tool chain}\par
  \texttt{timesfm2\_forecast}\par\vspace{4pt}
  
  \textbf{Sub-agent result}\par
  \texttt{answer = [236.906, 236.682, ..., 230.766]}\par\vspace{4pt}
  
  \textbf{Reasoning text}\par
  \small\raggedright
  \texttt{timesfm2\_forecast} was run on the original \texttt{134}-point hourly price series with horizon \texttt{33} and returned a full task-valid forecast array.\par
  
  The recorded explanation says this candidate was smooth and mildly downward-sloping, which matched the weak-trend, high-autocorrelation series better than an unchanged plateau.\par
  
  The trace also notes that the mixed article should be used only as a light contextual prior rather than the main driver of the forecast shape.\par
  \end{tcolorbox}
  \end{minipage}
  \hfill
  \begin{minipage}[t]{0.48\textwidth}
  \begin{tcolorbox}[
    colback=green!3,
    colframe=green!45!black,
    boxrule=0.5pt,
    arc=3pt,
    left=4pt, right=4pt, top=3pt, bottom=3pt,
    fontupper=\small\raggedright
  ]
  \textbf{Sub-agent 1}\par
  \textbf{Tool chain}\par
  \texttt{chronos2\_forecast}\par\vspace{4pt}
  
  \textbf{Sub-agent result}\par
  \texttt{answer = [237.000, 237.000, ..., 237.000]}\par\vspace{4pt}
  
  \textbf{Reasoning text}\par
  \small\raggedright
  \texttt{chronos2\_forecast} was run on the same original \texttt{134}-point hourly series with horizon \texttt{33} and returned a task-valid forecast array.\par
  
  The result was a flat constant path at \texttt{237.0} for all \texttt{33} steps, which the later learning analysis describes as underfitting the evolving hourly structure.\par
  
  This rigid plateau became the decisive failure signal when the two forecast candidates were evaluated against ground truth.
  \end{tcolorbox}
  \end{minipage}
  \vspace{4pt}
  \begin{minipage}[t]{1.00\textwidth}
  \begin{tcolorbox}[
    colback=purple!3,
    colframe=purple!45!black,
    boxrule=0.5pt,
    arc=3pt,
    left=4pt, right=4pt, top=3pt, bottom=3pt,
    fontupper=\small\raggedright
  ]
  \textbf{Ground truth Comparison and Final Learning Summary}\par\vspace{4pt}
  
  \textbf{Ground truth evaluation}\par
  \begin{quote}
  \small
  \texttt{sub-agent 0 / timesfm2\_forecast}: \texttt{MAE = 2.873}, \texttt{RMSE = 3.743}, \texttt{MAPE = 1.241}, \texttt{MSE = 14.012}\par
  \texttt{sub-agent 1 / chronos2\_forecast}: \texttt{MAE = 5.336}, \texttt{RMSE = 6.465}, \texttt{MAPE = 2.322}, \texttt{MSE = 41.792}
  \end{quote}
  
  \textbf{Final learning summary}\par
  \begin{quote}
  \small
  \texttt{insight}: The TimesFM 2.0 candidate was the stronger path for this \texttt{33}-step AAP hourly forecast. Its prediction captured a plausible mild downward drift and achieved materially lower error than the Chronos-2 constant-line candidate in post-hoc evaluation.\par\vspace{4pt}
  
  \texttt{recommendation}: For similar financial series with weak trend, high autocorrelation, and neutral-to-mild news context, prefer a foundation forecast path like \texttt{timesfm2\_forecast} and treat the news as a light contextual prior unless it is strongly directional or tightly event-timed.
  \end{quote}
  \end{tcolorbox}
  \end{minipage}
  \caption{Execution sample on finance forecast task.}
  \end{figure*}
  
  \begin{figure*}[t]
  \centering
  \begin{minipage}[t]{1.00\textwidth}
  \begin{tcolorbox}[
    colback=gray!10,
    colframe=gray!55,
    boxrule=0.5pt,
    arc=3pt,
    left=4pt, right=4pt, top=3pt, bottom=3pt,
    fontupper=\small\raggedright
  ]
  \textbf{Training Example 3}\par\vspace{4pt}
  
  \textbf{Task / Sample}\par
  \texttt{weather\_aligned\_long:USW00012815\_29.json:115:trend}\par
  Task: analyze the past 14 days of hourly temperature readings and predict the next 3-day temperature trend label.\par\vspace{4pt}
  
  \textbf{Task metadata and text excerpts used below}\par
  The trace records a weather long future-trend task from \texttt{2010-04-23 00:00:00} to \texttt{2010-05-06 23:00:00}. The article is end-aligned severe-weather text with thunderstorm and funnel-cloud context.\par
  \end{tcolorbox}
  \end{minipage}
  \vspace{4pt}
  \begin{minipage}[t]{1.00\textwidth}
  \begin{tcolorbox}[
    colback=blue!3,
    colframe=blue!45!black,
    boxrule=0.5pt,
    arc=3pt,
    left=4pt, right=4pt, top=3pt, bottom=3pt,
    fontupper=\small\raggedright
  ]
  \textbf{Main Agent}\par\vspace{4pt}
  
  \textbf{Two branch outcomes}\par
  Sub-agent 0 finished with \texttt{llm\_reasoning\_classification}.\par
  Sub-agent 1 finished with \texttt{chronos2\_forecast -> llm\_reasoning\_classification}.\par
  The trace later evaluated the two labels against the stored ground truth.\par
  \end{tcolorbox}
  \end{minipage}
  \vspace{4pt}
  \begin{minipage}[t]{0.37\textwidth}
  \begin{tcolorbox}[
    colback=green!3,
    colframe=green!45!black,
    boxrule=0.5pt,
    arc=3pt,
    left=4pt, right=4pt, top=3pt, bottom=3pt,
    fontupper=\small\raggedright
  ]
  \textbf{Sub-agent 0}\par
  \textbf{Tool chain}\par
  \texttt{llm\_reasoning\_classification}\par\vspace{4pt}
  
  \textbf{Sub-agent result}\par
  \texttt{answer = decreasing}\par\vspace{4pt}
  
  \textbf{Reasoning text}\par
  \small\raggedright
  The trace summary states that \texttt{llm\_reasoning\_classification} used the hourly temperature series together with the boundary storm narrative and the task's exact day-mean comparison rule.\par
  
  That tool returned \texttt{decreasing}.
  \end{tcolorbox}
  \end{minipage}
  \hfill
  \begin{minipage}[t]{0.61\textwidth}
  \begin{tcolorbox}[
    colback=red!3,
    colframe=red!45!black,
    boxrule=0.5pt,
    arc=3pt,
    left=4pt, right=4pt, top=3pt, bottom=3pt,
    fontupper=\small\raggedright
  ]
  \textbf{Sub-agent 1}\par
  \textbf{Tool chain}\par
  \texttt{chronos2\_forecast -> llm\_reasoning\_classification}\par\vspace{4pt}
  
  \textbf{Sub-agent result}\par
  \texttt{answer = increasing}\par\vspace{4pt}
  
  \textbf{Reasoning text}\par
  \small\raggedright
  \texttt{chronos2\_forecast} generated a 72-hour forecast.\par
  
  The recorded reasoning then used the exact comparison values
  \texttt{last observed 24-hour mean = 26.63125},
  \texttt{first predicted 24-hour mean = 29.75},
  and
  \texttt{difference = 3.11875}.\par
  
  Because that difference exceeded the prompt threshold, the branch returned \texttt{increasing}.
  \end{tcolorbox}
  \end{minipage}
  \vspace{4pt}
  \begin{minipage}[t]{1.00\textwidth}
  \begin{tcolorbox}[
    colback=purple!3,
    colframe=purple!45!black,
    boxrule=0.5pt,
    arc=3pt,
    left=4pt, right=4pt, top=3pt, bottom=3pt,
    fontupper=\small\raggedright
  ]
  \textbf{Ground truth Comparison and Final Learning Summary}\par\vspace{4pt}
  
  \textbf{Ground truth evaluation}\par
  \begin{quote}
  \small
  \texttt{sub-agent 0}: \texttt{decreasing} $\rightarrow$ \texttt{correct = true}\par
  \texttt{sub-agent 1}: \texttt{increasing} $\rightarrow$ \texttt{correct = false}\par
  \texttt{ground\_truth = decreasing}
  \end{quote}
  
  \textbf{Final learning summary}\par
  \begin{quote}
  \small
  \texttt{insight}: The successful path prioritized the last 24-hour regime and end-aligned thunderstorm/funnel-cloud context over a generic forecast model. The chronos2-based path overpredicted warming and missed the task's required boundary-sensitive day-mean comparison.\par\vspace{4pt}
  
  \texttt{recommendation}: When storm or severe-weather text occurs near the forecast boundary, compare at least one text-aware candidate against one forecast-model candidate, and favor the branch that respects the explicit 24-hour mean rule under regime-break conditions.
  \end{quote}
  \end{tcolorbox}
  \end{minipage}
  \caption{Execution sample on weather future-trend task.}
  \end{figure*}
  
  \begin{figure*}[t]
  \centering
  \begin{minipage}[t]{1.00\textwidth}
  \begin{tcolorbox}[
    colback=gray!10,
    colframe=gray!55,
    boxrule=0.5pt,
    arc=3pt,
    left=4pt, right=4pt, top=3pt, bottom=3pt,
    fontupper=\small\raggedright
  ]
  \textbf{Training Example 4}\par\vspace{4pt}
  
  \textbf{Task / Sample}\par
  \texttt{finance\_aligned\_short:00400.parquet\_row:281:trend}\par
  Task: predict the future return bucket from an aligned short finance series and article.\par\vspace{4pt}
  
  \textbf{Task metadata and text excerpts used below}\par
  The trace records a finance short trend task. The news is an earnings-preview article with strong year-over-year growth expectations but an inconclusive earnings-beat signal.\par
  \end{tcolorbox}
  \end{minipage}
  \vspace{4pt}
  \begin{minipage}[t]{1.00\textwidth}
  \begin{tcolorbox}[
    colback=blue!3,
    colframe=blue!45!black,
    boxrule=0.5pt,
    arc=3pt,
    left=4pt, right=4pt, top=3pt, bottom=3pt,
    fontupper=\small\raggedright
  ]
  \textbf{Main Agent}\par\vspace{4pt}
  
  \textbf{Two branch outcomes}\par
  Sub-agent 0 finished with \texttt{ets\_forecast -> value\_at -> value\_at -> llm\_reasoning\_classification}.\par
  Sub-agent 1 finished with \texttt{moirai1\_1\_r\_base\_forecast -> llm\_reasoning\_classification}.\par
  The trace later compared these two bucket labels against the stored ground truth.\par
  \end{tcolorbox}
  \end{minipage}
  \vspace{4pt}
  \begin{minipage}[t]{0.58\textwidth}
  \begin{tcolorbox}[
    colback=green!3,
    colframe=green!45!black,
    boxrule=0.5pt,
    arc=3pt,
    left=4pt, right=4pt, top=3pt, bottom=3pt,
    fontupper=\small\raggedright
  ]
  \textbf{Sub-agent 0}\par
  \textbf{Tool chain}\par
  \texttt{ets\_forecast -> value\_at -> value\_at -> llm\_reasoning\_classification}\par\vspace{4pt}
  
  \textbf{Sub-agent result}\par
  \texttt{answer = -2\% \textasciitilde +2\%}\par\vspace{4pt}
  
  \textbf{Reasoning text}\par
  \small\raggedright
  \texttt{ets\_forecast} produced a 78-step forecast ending at \texttt{62.544}.\par
  
  \texttt{value\_at} on \texttt{original\_input} retrieved the latest observed price as \texttt{61.94}.\par
  
  The second \texttt{value\_at} computed \texttt{pct\_change\_vs\_reference = +0.976\%}, which stayed below the \texttt{+2\%} cutoff, so the branch returned \texttt{-2\% \textasciitilde +2\%}.
  \end{tcolorbox}
  \end{minipage}
  \hfill
  \begin{minipage}[t]{0.40\textwidth}
  \begin{tcolorbox}[
    colback=red!3,
    colframe=red!45!black,
    boxrule=0.5pt,
    arc=3pt,
    left=4pt, right=4pt, top=3pt, bottom=3pt,
    fontupper=\small\raggedright
  ]
  \textbf{Sub-agent 1}\par
  \textbf{Tool chain}\par
  \texttt{moirai1\_1\_r\_base\_forecast -> llm\_reasoning\_classification}\par\vspace{4pt}
  
  \textbf{Sub-agent result}\par
  \texttt{answer = +2\% \textasciitilde +4\%}\par\vspace{4pt}
  
  \textbf{Reasoning text}\par
  \small\raggedright
  \texttt{moirai1\_1\_r\_base\_forecast} produced a rising 96-step path ending at \texttt{64.382}.\par
  
  The recorded reasoning states that this implied an approximate forward return of \texttt{+3.94\%}, which mapped to \texttt{+2\% \textasciitilde +4\%}.
  \end{tcolorbox}
  \end{minipage}
  \vspace{4pt}
  \begin{minipage}[t]{1.00\textwidth}
  \begin{tcolorbox}[
    colback=purple!3,
    colframe=purple!45!black,
    boxrule=0.5pt,
    arc=3pt,
    left=4pt, right=4pt, top=3pt, bottom=3pt,
    fontupper=\small\raggedright
  ]
  \textbf{Ground truth Comparison and Final Learning Summary}\par\vspace{4pt}
  
  \textbf{Ground truth evaluation}\par
  \begin{quote}
  \small
  \texttt{sub-agent 0}: \texttt{-2\% \textasciitilde +2\%} $\rightarrow$ \texttt{correct = true}\par
  \texttt{sub-agent 1}: \texttt{+2\% \textasciitilde +4\%} $\rightarrow$ \texttt{correct = false}\par
  \texttt{ground\_truth = -2\% \textasciitilde +2\%}
  \end{quote}
  
  \textbf{Final learning summary}\par
  \begin{quote}
  \small
  \texttt{insight}: The better path was the forecast-to-endpoint workflow from \texttt{ets\_forecast -> value\_at -> llm\_reasoning\_classification}. It computed the latest observed price at \texttt{61.94} and a forecast endpoint at \texttt{62.544}, implying only about \texttt{+0.98\%}. The alternative branch projected about \texttt{+3.94\%} and overcalled the move.\par\vspace{4pt}
  
  \texttt{recommendation}: When weak or routine earnings-preview news accompanies a highly autocorrelated intraday series, prefer an explicit endpoint-based bucket decision from a concrete forecast path and compare the endpoint directly with the latest observed price.
  \end{quote}
  \end{tcolorbox}
  \end{minipage}
  \caption{Execution sample on finance trend task.}
  \end{figure*}
  
  \begin{figure*}[t]
  \centering
  \begin{minipage}[t]{1.00\textwidth}
  \begin{tcolorbox}[
    colback=gray!10,
    colframe=gray!55,
    boxrule=0.5pt,
    arc=3pt,
    left=4pt, right=4pt, top=3pt, bottom=3pt,
    fontupper=\small\raggedright
  ]
  \textbf{Training Example 5}\par\vspace{4pt}
  
  \textbf{Task / Sample}\par
  \texttt{weather\_aligned\_long:USW00003947\_17.json:73:trend\_past}\par
  Task: classify the temperature trend over the observed past 14 days.\par\vspace{4pt}
  
  \textbf{Task metadata and text excerpts used below}\par
  The trace records a weather long past-trend task on hourly temperatures from \texttt{2017-07-09 00:00:00} to \texttt{2017-07-22 23:00:00}. The aligned text is a severe-weather narrative with hail and thunderstorm wind near the end of the window.\par
  \end{tcolorbox}
  \end{minipage}
  \vspace{4pt}
  \begin{minipage}[t]{1.00\textwidth}
  \begin{tcolorbox}[
    colback=blue!3,
    colframe=blue!45!black,
    boxrule=0.5pt,
    arc=3pt,
    left=4pt, right=4pt, top=3pt, bottom=3pt,
    fontupper=\small\raggedright
  ]
  \textbf{Main Agent}\par\vspace{4pt}
  
  \textbf{Two branch outcomes}\par
  Sub-agent 0 finished with \texttt{llm\_reasoning\_classification}.\par
  Sub-agent 1 finished with \texttt{temporal\_align\_text -> llm\_reasoning\_classification}.\par
  The trace later evaluated both labels against the stored ground truth.\par
  \end{tcolorbox}
  \end{minipage}
  \vspace{4pt}
  \begin{minipage}[t]{0.49\textwidth}
  \begin{tcolorbox}[
    colback=green!3,
    colframe=green!45!black,
    boxrule=0.5pt,
    arc=3pt,
    left=4pt, right=4pt, top=3pt, bottom=3pt,
    fontupper=\small\raggedright
  ]
  \textbf{Sub-agent 0}\par
  \textbf{Tool chain}\par
  \texttt{llm\_reasoning\_classification}\par\vspace{4pt}
  
  \textbf{Sub-agent result}\par
  \texttt{answer = stable}\par\vspace{4pt}
  
  \textbf{Reasoning text}\par
  \small\raggedright
  The trace says this branch was instructed to treat every 24 hourly readings as one day, compute 14 daily mean temperatures, fit a linear trend on those daily means, and apply the prompt thresholds.\par
  
  That branch used the storm article only as secondary context and returned \texttt{stable}.
  \end{tcolorbox}
  \end{minipage}
  \hfill
  \begin{minipage}[t]{0.49\textwidth}
  \begin{tcolorbox}[
    colback=red!3,
    colframe=red!45!black,
    boxrule=0.5pt,
    arc=3pt,
    left=4pt, right=4pt, top=3pt, bottom=3pt,
    fontupper=\small\raggedright
  ]
  \textbf{Sub-agent 1}\par
  \textbf{Tool chain}\par
  \texttt{temporal\_align\_text -> llm\_reasoning\_classification}\par\vspace{4pt}
  
  \textbf{Sub-agent result}\par
  \texttt{answer = increasing}\par\vspace{4pt}
  
  \textbf{Reasoning text}\par
  \small\raggedright
  \texttt{temporal\_align\_text} aligned the severe-weather article near the boundary.\par
  
  The same execution trace records that the classifier still used the daily-mean slope rule, but ultimately returned \texttt{increasing}.\par
  \end{tcolorbox}
  \end{minipage}
  \vspace{4pt}
  \begin{minipage}[t]{1.00\textwidth}
  \begin{tcolorbox}[
    colback=purple!3,
    colframe=purple!45!black,
    boxrule=0.5pt,
    arc=3pt,
    left=4pt, right=4pt, top=3pt, bottom=3pt,
    fontupper=\small\raggedright
  ]
  \textbf{Ground truth Comparison and Final Learning Summary}\par\vspace{4pt}
  
  \textbf{Ground truth evaluation}\par
  \begin{quote}
  \small
  \texttt{sub-agent 0}: \texttt{stable} $\rightarrow$ \texttt{correct = true}\par
  \texttt{sub-agent 1}: \texttt{increasing} $\rightarrow$ \texttt{correct = false}\par
  \texttt{ground\_truth = stable}
  \end{quote}
  
  \textbf{Final learning summary}\par
  \begin{quote}
  \small
  \texttt{insight}: For 14-day hourly weather trend tasks, the reliable path is to classify from the observed window's 14 daily mean temperatures using the prompt thresholds. In this run, the candidate that returned \texttt{stable} was correct, while a text-aligned candidate drifted to \texttt{increasing} and failed.\par\vspace{4pt}
  
  \texttt{recommendation}: Prefer a day-aligned daily-mean trend workflow over multimodal narrative influence, and treat storm text only as secondary context unless it visibly changes the whole-window daily-mean trend.
  \end{quote}
  \end{tcolorbox}
  \end{minipage}
  \caption{Execution sample on weather past-trend task.}
  \end{figure*}
  
  \subsection{Inference Examples}
  
  These inference examples are reconstructed from row-level inference artifacts rather than from post-hoc summaries. Each case shows the injected memory rules, the realized tool chain, the stored step records, and the final \texttt{execution\_context}. The emphasis is on execution-side evidence reuse at test time: instead of describing the policy at a high level, we show exactly what memory was injected, which tools were called, what intermediate outputs were produced, and how the final prediction was grounded in those stored artifacts.
  
  \begin{figure*}[t]
  \centering
  \begin{minipage}[t]{1.00\textwidth}
  \begin{tcolorbox}[
    colback=gray!10,
    colframe=gray!55,
    boxrule=0.5pt,
    arc=3pt,
    left=4pt, right=4pt, top=3pt, bottom=3pt,
    fontupper=\small\raggedright
  ]
  \textbf{Inference Example 1}\par\vspace{4pt}
  
  \textbf{Task / Sample}\par
  \texttt{weather\_aligned\_long:USW00003017\_11.json:3:forecast}\par
  Task: predict the next 72 hourly temperature values from a 14-day hourly weather series and aligned severe-weather text.\par\vspace{4pt}
  
  \textbf{Task metadata and text excerpts used below}\par
  The row artifact records a weather long forecast task at station \texttt{USW00003017}. The history shows a strong repeated \texttt{24}-hour rhythm, and the aligned text describes hail and thunderstorm-wind events near the forecast boundary.\par
  \end{tcolorbox}
  \end{minipage}
  \vspace{4pt}
  \begin{minipage}[t]{1.00\textwidth}
  \begin{tcolorbox}[
    colback=orange!3,
    colframe=orange!50!black,
    boxrule=0.5pt,
    arc=3pt,
    left=4pt, right=4pt, top=3pt, bottom=3pt,
    fontupper=\small\raggedright
  ]
  \textbf{Inference Flow}\par\vspace{4pt}
  
  \textbf{Injected skills text}\par
  \begin{quote}
  \small
  When the daily cycle is strong but the latest endpoint looks unusually warm or storm-distorted, compare it against the recent repeating temperature band, soften the start back toward that band, and prefer \texttt{sundial\_base\_128m\_forecast}.
  \end{quote}\vspace{4pt}

  \textbf{Injected memory text}\par
  \begin{quote}
  \small
  1. Forecasts should preserve a believable \texttt{24}-hour temperature rhythm across all \texttt{72} hourly steps.\par
  2. Treat storm text as a short-lived modifier unless it clearly signals a real regime break such as a cold front or snow.\par
  3. Start the forecast close to the last observed temperature before continuing the next diurnal cycle.
  \end{quote}\vspace{4pt}
  
  \textbf{Tool chain}\par
  \texttt{sundial\_base\_128m\_forecast -> timesfm2\_forecast}\par\vspace{4pt}
  
  \textbf{Step records}\par
  \begin{quote}
  \small
  1. \texttt{sundial\_base\_128m\_forecast}: produced a smooth \texttt{72}-step temperature path that preserved overnight cooling and daytime warming.\par\vspace{4pt}
  
  2. \texttt{timesfm2\_forecast}: produced a second real \texttt{72}-step forecast candidate for the same horizon.\par\vspace{4pt}
  
  3. Final prediction metrics in the row artifact: \texttt{MAE = 1.687}, \texttt{RMSE = 2.323}, \texttt{MAPE = 8.090}, \texttt{MSE = 5.398}.
  \end{quote}\vspace{4pt}
  
  \textbf{execution\_context field}\par
  \begin{quote}
  \small
  The historical series shows a strong repeated \texttt{24}-hour temperature rhythm with only a brief storm event near the boundary.\par
  \texttt{sundial\_base\_128m\_forecast} preserved smooth overnight cooling and daytime warming while softening the warm endpoint back toward the recent band.\par
  \texttt{timesfm2\_forecast} was plausible, but \texttt{sundial\_base\_128m\_forecast} aligned better with the sample guidance for a storm-distorted boundary.
  \end{quote}\vspace{4pt}
  
  \textbf{Observed outcome}\par
  Prediction: \texttt{[27.142, 25.426, ..., 31.569]}\par
  Ground truth: \texttt{[26.100, 25.000, ..., 24.300]}
  \end{tcolorbox}
  \end{minipage}
  \caption{Inference sample on weather forecast task.}
  \end{figure*}
  
  \begin{figure*}[t]
  \centering
  \begin{minipage}[t]{1.00\textwidth}
  \begin{tcolorbox}[
    colback=gray!10,
    colframe=gray!55,
    boxrule=0.5pt,
    arc=3pt,
    left=4pt, right=4pt, top=3pt, bottom=3pt,
    fontupper=\small\raggedright
  ]
  \textbf{Inference Example 2}\par\vspace{4pt}
  
  \textbf{Task / Sample}\par
  \texttt{finance\_aligned\_long:00112.parquet\_row:112:trend}\par
  Task: predict the finance long-horizon future return bucket.\par\vspace{4pt}
  
  \textbf{Task metadata and text excerpts used below}\par
  The row artifact records a finance long trend task. The execution used endpoint evidence plus news-event checking rather than a full-month forecast-first path.\par
  \end{tcolorbox}
  \end{minipage}
  \vspace{4pt}
  \begin{minipage}[t]{1.00\textwidth}
  \begin{tcolorbox}[
    colback=orange!3,
    colframe=orange!50!black,
    boxrule=0.5pt,
    arc=3pt,
    left=4pt, right=4pt, top=3pt, bottom=3pt,
    fontupper=\small\raggedright
  ]
  \textbf{Inference Flow}\par\vspace{4pt}
  
  \textbf{Injected skills text}\par
  \begin{quote}
  \small
  Judge the bucket from the latest regime versus the current endpoint rather than the full-month slope; anchor on the latest level with \texttt{value\_at}, use \texttt{detect\_trend} only to summarize the recent tail, and keep the older month drift secondary.
  \end{quote}\vspace{4pt}

  \textbf{Injected memory text}\par
  \begin{quote}
  \small
  1. Weight the last regime and endpoint behavior more than the full-month trend.\par
  2. Match the chosen bucket to the implied percentage move from the latest price.\par
  3. Treat routine bullish commentary as weaker magnitude evidence than a fresh company catalyst.
  \end{quote}\vspace{4pt}
  
  \textbf{Tool chain}\par
  \texttt{value\_at -> text\_to\_event -> llm\_reasoning\_classification}\par\vspace{4pt}
  
  \textbf{Step records}\par
  \begin{quote}
  \small
  1. \texttt{value\_at}: latest observed price \texttt{= 19.375}, with \texttt{pct\_change\_vs\_reference = 35.55\%} from the first observed price.\par\vspace{4pt}
  
  2. \texttt{text\_to\_event}: \texttt{events = []}, \texttt{n\_events = 0}.\par\vspace{4pt}
  
  3. \texttt{llm\_reasoning\_classification}: \texttt{answer = +2\% \textasciitilde +4\%}.
  \end{quote}\vspace{4pt}
  
  \textbf{execution\_context field}\par
  \begin{quote}
  \small
  The latest observed price is \texttt{19.375}, well above the start of the month, confirming strong upward momentum into the endpoint.\par
  The series diagnostics indicate a significant uptrend but also non-stationarity, so the latest regime matters more than the full-month average drift.\par
  News is supportive of fundamentals, but \texttt{text\_to\_event} found no strong discrete catalyst, which argues for a moderate bullish bucket instead of the most extreme one.
  \end{quote}\vspace{4pt}
  
  \textbf{Observed outcome}\par
  Prediction: \texttt{+2\% \textasciitilde +4\%}\par
  Ground truth: \texttt{+2\% \textasciitilde +4\%}
  \end{tcolorbox}
  \end{minipage}
  \caption{Inference sample on finance trend task.}
  \end{figure*}
  
  \begin{figure*}[t]
  \centering
  \begin{minipage}[t]{1.00\textwidth}
  \begin{tcolorbox}[
    colback=gray!10,
    colframe=gray!55,
    boxrule=0.5pt,
    arc=3pt,
    left=4pt, right=4pt, top=3pt, bottom=3pt,
    fontupper=\small\raggedright
  ]
  \textbf{Inference Example 3}\par\vspace{4pt}
  
  \textbf{Task / Sample}\par
  \texttt{weather\_aligned\_long:USW00003017\_16.json:8:trend}\par
  Task: predict the next 3-day weather trend label from a 14-day hourly series and aligned weather text.\par\vspace{4pt}
  
  \textbf{Task metadata and text excerpts used below}\par
  The row artifact records a weather long future-trend task. The execution emphasized daily segmentation and last-day mean measurement before final classification.\par
  \end{tcolorbox}
  \end{minipage}
  \vspace{4pt}
  \begin{minipage}[t]{1.00\textwidth}
  \begin{tcolorbox}[
    colback=orange!3,
    colframe=orange!50!black,
    boxrule=0.5pt,
    arc=3pt,
    left=4pt, right=4pt, top=3pt, bottom=3pt,
    fontupper=\small\raggedright
  ]
  \textbf{Inference Flow}\par\vspace{4pt}
  
  \textbf{Injected skills text}\par
  \begin{quote}
  \small
  Compare the mean of the first future \texttt{24} hours against the mean of the last observed \texttt{24} hours using the exact \texttt{\(\pm 0.5\)} rule, and prefer exact daily segmentation plus boundary-aware next-day reasoning over general slope impressions.
  \end{quote}\vspace{4pt}

  \textbf{Injected memory text}\par
  \begin{quote}
  \small
  1. Classify from the mean of the last observed 24 hours versus the mean of the first predicted 24 hours using the exact \texttt{\(\pm 0.5\)} thresholds.\par
  2. Prioritize the latest 24-hour regime and any end-aligned storm context over the full 14-day slope.\par
  3. Segment the series into exact daily windows before reasoning when the task is explicitly about comparing daily means.
  \end{quote}\vspace{4pt}
  
  \textbf{Tool chain}\par
  \texttt{segment -> window\_stats -> llm\_reasoning\_classification}\par\vspace{4pt}
  
  \textbf{Step records}\par
  \begin{quote}
  \small
  1. \texttt{segment}: the \texttt{336} hourly values were divided into \texttt{14} exact daily windows.\par\vspace{4pt}
  
  2. \texttt{window\_stats}: on the last observed 24 hours, \texttt{mean = 16.435416666666665}.\par\vspace{4pt}
  
  3. \texttt{llm\_reasoning\_classification}: \texttt{answer = increasing}.
  \end{quote}\vspace{4pt}
  
  \textbf{execution\_context field}\par
  \begin{quote}
  \small
  The \texttt{336} hourly values divide cleanly into \texttt{14} exact 24-hour daily windows, so a like-for-like daily comparison is appropriate.\par
  \texttt{window\_stats} on the last observed 24 hours returned a mean of \texttt{16.435} for the reference day.\par
  \texttt{llm\_reasoning\_classification} used that reference mean plus the end-aligned severe-weather context and returned \texttt{increasing}.
  \end{quote}\vspace{4pt}
  
  \textbf{Observed outcome}\par
  Prediction: \texttt{increasing}\par
  Ground truth: \texttt{increasing}
  \end{tcolorbox}
  \end{minipage}
  \caption{Inference sample on weather future-trend task.}
  \end{figure*}
  
  \begin{figure*}[t]
  \centering
  \begin{minipage}[t]{1.00\textwidth}
  \begin{tcolorbox}[
    colback=gray!10,
    colframe=gray!55,
    boxrule=0.5pt,
    arc=3pt,
    left=4pt, right=4pt, top=3pt, bottom=3pt,
    fontupper=\small\raggedright
  ]
  \textbf{Inference Example 4}\par\vspace{4pt}
  
  \textbf{Task / Sample}\par
  \texttt{weather\_aligned\_short:USW00003017\_11.json:3:trend}\par
  Task: predict the next-1-day weather trend label from a 7-day hourly series and aligned weather text.\par\vspace{4pt}
  
  \textbf{Task metadata and text excerpts used below}\par
  The row artifact records a weather short future-trend task. This example contains both a primary forecast path and an additional challenger forecast check before the final answer.\par
  \end{tcolorbox}
  \end{minipage}
  \vspace{4pt}
  \begin{minipage}[t]{1.00\textwidth}
  \begin{tcolorbox}[
    colback=orange!3,
    colframe=orange!50!black,
    boxrule=0.5pt,
    arc=3pt,
    left=4pt, right=4pt, top=3pt, bottom=3pt,
    fontupper=\small\raggedright
  ]
  \textbf{Inference Flow}\par\vspace{4pt}
  
  \textbf{Injected skills text}\par
  \begin{quote}
  \small
  Compute the mean of the final observed \texttt{24}-hour window, generate an explicit next-day \texttt{24}-step forecast, and classify only from the difference between those two means instead of from the weekly slope or weather narrative alone.
  \end{quote}\vspace{4pt}

  \textbf{Injected memory text}\par
  \begin{quote}
  \small
  1. Base the label on the difference between the observed last-day mean and the first forecast-day mean rather than on the full-series slope.\par
  2. Treat compressed last-day windows near the \texttt{\(\pm 0.5\)} cutoff as threshold-sensitive cases.\par
  3. When boundary weather makes the mean shift near the cutoff, compare more than one plausible next-day forecast scenario before classifying.
  \end{quote}\vspace{4pt}
  
  \textbf{Tool chain}\par
  \texttt{segment -> chronos2\_forecast -> window\_stats -> window\_stats -> aurora\_forecast -> llm\_reasoning\_classification}\par\vspace{4pt}
  
  \textbf{Step records}\par
  \begin{quote}
  \small
  1. \texttt{window\_stats} on the last observed 24-hour window gave \texttt{mean = 20.239583333333332}.\par\vspace{4pt}
  
  2. \texttt{window\_stats} on the \texttt{chronos2\_forecast} output gave \texttt{mean = 24.432291666666668} and \texttt{mean\_delta\_vs\_reference = 4.192291666666669}.\par\vspace{4pt}
  
  3. \texttt{aurora\_forecast} was executed as an additional boundary-aware challenger forecast.\par\vspace{4pt}
  
  4. \texttt{llm\_reasoning\_classification}: \texttt{answer = increasing}.
  \end{quote}\vspace{4pt}
  
  \textbf{execution\_context field}\par
  \begin{quote}
  \small
  The last observed 24-hour mean from \texttt{window\_stats} on \texttt{original\_input} is \texttt{20.24}.\par
  \texttt{window\_stats} on the \texttt{chronos2\_forecast} output gave a first forecast-day mean of \texttt{24.432}, a \texttt{+4.192} increase versus the last observed day.\par
  Both primary and challenger boundary-aware forecast paths indicate the next day's mean temperature is well above the last observed day's mean, exceeding the \texttt{+0.5} threshold.
  \end{quote}\vspace{4pt}
  
  \textbf{Observed outcome}\par
  Prediction: \texttt{increasing}\par
  Ground truth: \texttt{increasing}
  \end{tcolorbox}
  \end{minipage}
  \caption{Inference sample on weather future-trend task.}
  \end{figure*}
  
  \begin{figure*}[t]
  \centering
  \begin{minipage}[t]{1.00\textwidth}
  \begin{tcolorbox}[
    colback=gray!10,
    colframe=gray!55,
    boxrule=0.5pt,
    arc=3pt,
    left=4pt, right=4pt, top=3pt, bottom=3pt,
    fontupper=\small\raggedright
  ]
  \textbf{Inference Example 5}\par\vspace{4pt}
  
  \textbf{Task / Sample}\par
  \texttt{finance\_aligned\_short:00100.parquet\_row:100:forecast}\par
  Task: predict the next 78 intraday stock prices from a short-horizon finance series and aligned finance text.\par\vspace{4pt}
  
  \textbf{Task metadata and text excerpts used below}\par
  The row artifact records a finance short forecast task. The aligned text is generic Nasdaq volume reporting, so the execution emphasizes recent price structure rather than extracting a strong textual catalyst.\par
  \end{tcolorbox}
  \end{minipage}
  \vspace{4pt}
  \begin{minipage}[t]{1.00\textwidth}
  \begin{tcolorbox}[
    colback=orange!3,
    colframe=orange!50!black,
    boxrule=0.5pt,
    arc=3pt,
    left=4pt, right=4pt, top=3pt, bottom=3pt,
    fontupper=\small\raggedright
  ]
  \textbf{Inference Flow}\par\vspace{4pt}
  
  \textbf{Injected skills text}\par
  \begin{quote}
  \small
  When recent prices are highly autocorrelated and move through many small continuous changes while the text is weak or generic, preserve dense local drift, downweight the text, and prefer \texttt{timesfm2\_forecast}.
  \end{quote}\vspace{4pt}

  \textbf{Injected memory text}\par
  \begin{quote}
  \small
  1. Use aligned news as a light directional calibration unless it contains a clear near-horizon catalyst.\par
  2. Return exactly the prompt-specified number of float prices in the required format.\par
  3. Prefer \texttt{timesfm2\_forecast} when a \texttt{5}-minute financial series needs smooth fine-grained continuation rather than quantized level holding.
  \end{quote}\vspace{4pt}
  
  \textbf{Tool chain}\par
  \texttt{timesfm2\_forecast -> moirai1\_1\_r\_base\_forecast}\par\vspace{4pt}
  
  \textbf{Step records}\par
  \begin{quote}
  \small
  1. \texttt{timesfm2\_forecast}: generated a smooth \texttt{78}-step path clustered near \texttt{51.41} and slowly easing lower.\par\vspace{4pt}
  
  2. \texttt{moirai1\_1\_r\_base\_forecast}: generated a second \texttt{78}-step candidate with repeated oscillations and sharper spikes.\par\vspace{4pt}
  
  3. Final prediction metrics in the row artifact: \texttt{MAE = 0.190}, \texttt{RMSE = 0.219}, \texttt{MAPE = 0.371}, \texttt{MSE = 0.048}.
  \end{quote}\vspace{4pt}
  
  \textbf{execution\_context field}\par
  \begin{quote}
  \small
  \texttt{timesfm2\_forecast} generated a smooth \texttt{78}-step path clustered near \texttt{51.41} and slowly easing lower.\par
  \texttt{moirai1\_1\_r\_base\_forecast} generated a much choppier \texttt{78}-step path with repeated sharp oscillations and spikes.\par
  The news text was generic Nasdaq volume reporting, so recent price structure is the main forecast signal.
  \end{quote}\vspace{4pt}
  
  \textbf{Observed outcome}\par
  Prediction: \texttt{[51.414, 51.423, ..., 51.327]}\par
  Ground truth: \texttt{[51.367, 51.267, ..., 51.530]}
  \end{tcolorbox}
  \end{minipage}
  \caption{Inference sample on finance forecast task.}
  \end{figure*}

\section{Limitations and Broader Impacts}
\label{app:limitations_broader}

\subsection{Limitations}
\label{app:limitations}

TimeClaw is evaluated on a fixed 17-task MTBench suite covering finance and weather, so the empirical claims in this paper are limited to benchmarked, verifiable time-series tasks with the tool interfaces considered here. The method also depends on a frozen base LLM, an externally served API stack, and a predefined tool library, so performance may change under different backbones, tool sets, or deployment constraints.

Our learning protocol is offline rather than continual: we use a disjoint exploration corpus, fix an exploration budget of 300 samples per task, and then reuse distilled experience at inference time without online adaptation. We do not study sensitivity to exploration budget, domain shift beyond the MTBench-style setting, or real-world deployment under continuously changing data distributions.

Exploratory learning also introduces additional compute cost and a high volume of LLM API requests. In the current setup, a full exploration run takes about one day on two NVIDIA RTX 3090 GPUs and consumes a substantial number of tokens due to the generation of multiple candidate paths for each sample. Consequently, scaling to larger tool libraries or broader task suites will linearly increase both compute time and API expenditure. Finally, although we report benchmark metrics across all tasks, we do not include confidence intervals or formal significance tests, so small differences between close baselines should be interpreted cautiously.

\subsection{Broader Impacts}
\label{app:broader_impacts}

TimeClaw is intended for scientific and decision-support settings in which an agent must reason over time series, text, and tool outputs in an auditable way. Potential positive impacts include more transparent forecasting and reasoning workflows in domains such as weather analysis and financial monitoring, where reusable execution traces may help users inspect why a prediction was made.

Potential negative impacts also exist. In finance, overconfident use of agent forecasts could encourage inappropriate investment decisions; in weather or operational monitoring, incorrect predictions or over-trust in exploratory memories could mislead downstream action. Because TimeClaw improves agentic tool use rather than removing uncertainty, we view it as a decision-support framework rather than a substitute for domain experts, and its outputs should be used with human oversight and task-appropriate safeguards.

\subsection{The Use of Large Language Models (LLMs)}
\label{LLM-Usage}

We acknowledge the use of large language models (LLMs) exclusively for linguistic polishing and manuscript preparation. The application of AI was strictly limited to grammatical refinement, lexical enhancement, and structural editing to ensure clarity and professional presentation. All core scientific components—including the research methodology, experimental framework, data analysis, and technical derivations—represent the authors' original intellectual contributions. No substantive findings, data interpretations, or scientific insights were generated or influenced by AI tools.

\end{document}